\documentclass[]{style}
\usepackage[toc,page,header]{appendix}
\usepackage{wrapfig}

\usepackage{natbib}

\usepackage{CJKutf8}

\usepackage{xargs}

\usepackage{todonotes}

\usepackage{multirow}

\usepackage{cleveref}

\usepackage{amsmath}
\usepackage{amssymb}
\usepackage{mathtools}
\usepackage{amsthm}
\usepackage{dsfont}

\usepackage{svg}

\usepackage{mathrsfs}
\usepackage{adjustbox}
\usepackage{multicol}
\usepackage{tcolorbox}
\usepackage{changepage}
\usepackage{enumitem}
\usepackage{graphicx}
\usepackage{xcolor}
\usepackage{float}
\usepackage{threeparttable}
\usepackage{subcaption}
\usepackage{algorithm}
\usepackage{algpseudocode}
\usepackage{wrapfig}
\usepackage[table]{xcolor}
\usepackage{colortbl}
\usepackage{tabularx}
\usepackage{makecell}
\usepackage[dvipsnames]{xcolor}

\newcolumntype{g}{>{\columncolor{gray!10}}c}

\newcommand{\nameofmethod}{IB-Adapter}

\theoremstyle{plain}
\newtheorem{theorem}{Theorem}[section]
\newtheorem{proposition}[theorem]{Proposition}

\theoremstyle{definition}

\theoremstyle{remark}

\definecolor{catgray}{gray}{0.9}
\definecolor{skyblue}{rgb}{0.53,0.81,0.92}
\colorlet{skyblue!30}{skyblue!30!white}
\definecolor{customblue}{RGB}{70,130,180}

\renewcommand{\emph}[1]{\textit{#1}}

\usepackage{minitoc}
\usepackage{url}

\PassOptionsToPackage{table,xcdraw}{xcolor}
\usepackage{titletoc}
\usepackage{placeins}
\usepackage{pifont}

\definecolor{RowBlue}{HTML}{E9F2FB}
\definecolor{RowRed}{HTML}{F9EAEA}
\definecolor{Top1}{HTML}{50DB4B}
\definecolor{Top2}{HTML}{A5FFA2}
\definecolor{Top3}{HTML}{D9FFD9}
\definecolor{Sub1}{HTML}{EAB8B8}
\definecolor{Sub2}{HTML}{E4E4E4}

\title{StableVLA: Towards Robust Vision-Language-Action Models without Extra Data}

\author[1]{Yiyang Fu}
\author[2,3]{Chubin Zhang}
\author[1]{Shukai Gong}
\author[1]{Yufan Deng}
\author[4]{Kaiwei Sun}
\author[]{Qiyang Min}
\author[5]{Qibin Hou}
\author[2]{Yansong Tang}
\author[3\dagger]{Jianan Wang}
\author[1\dagger\dagger]{Daquan Zhou}

\affiliation[1]{Peking University}
\affiliation[2]{Tsinghua University}
\affiliation[3]{Astribot}
\affiliation[4]{Nanjing University}
\affiliation[5]{Nankai University}
\contribution[\dagger]{Project Leader}
\contribution[\dagger\dagger]{Corresponding author}

\abstract{
It is infeasible to encompass all possible disturbances within the training dataset. 
This raises a critical question regarding the robustness of Vision-Language-Action (VLA) models when encountering unseen real-world visual disturbances, particularly under imperfect visual conditions. 
In this work, we conduct a systematic study based on recent state-of-the-art VLA models and reveal a significant performance drop when visual disturbances absent from the training data are introduced. 
To mitigate this issue, we propose a lightweight adapter module grounded in information theory, termed the Information Bottleneck Adapter (\nameofmethod), which selectively filters potential noise from visual inputs. Without requiring any extra data or augmentation strategies, \nameofmethod~consistently improves over the baseline by an average of 30\%, while adding fewer than 10M parameters, demonstrating notable efficiency and effectiveness. Furthermore, even with a 14$\times$ smaller backbone (0.5B parameters) and no pre-training on the Open X-Embodiment dataset, our model StableVLA achieves robustness competitive with 7B-scale state-of-the-art VLAs. With negligible parameter overhead ($<$10M), our approach maintains accuracy on long-horizon tasks and surpasses OpenPi under both synthetic and physical visual corruptions.
}

\checkdata[
\raisebox{-0.2em}{\includegraphics[width=0.025\linewidth]{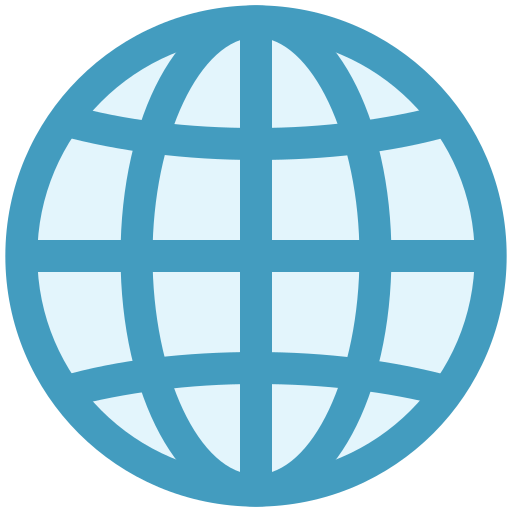}}~~Project Page]{\href{https://dagroup-pku.github.io/StableVLA/}{\texttt{https://dagroup-pku.github.io/StableVLA/}}
\\[-1.5ex]}

\checkdata[
\raisebox{-0.2em}{\includegraphics[width=0.025\linewidth]{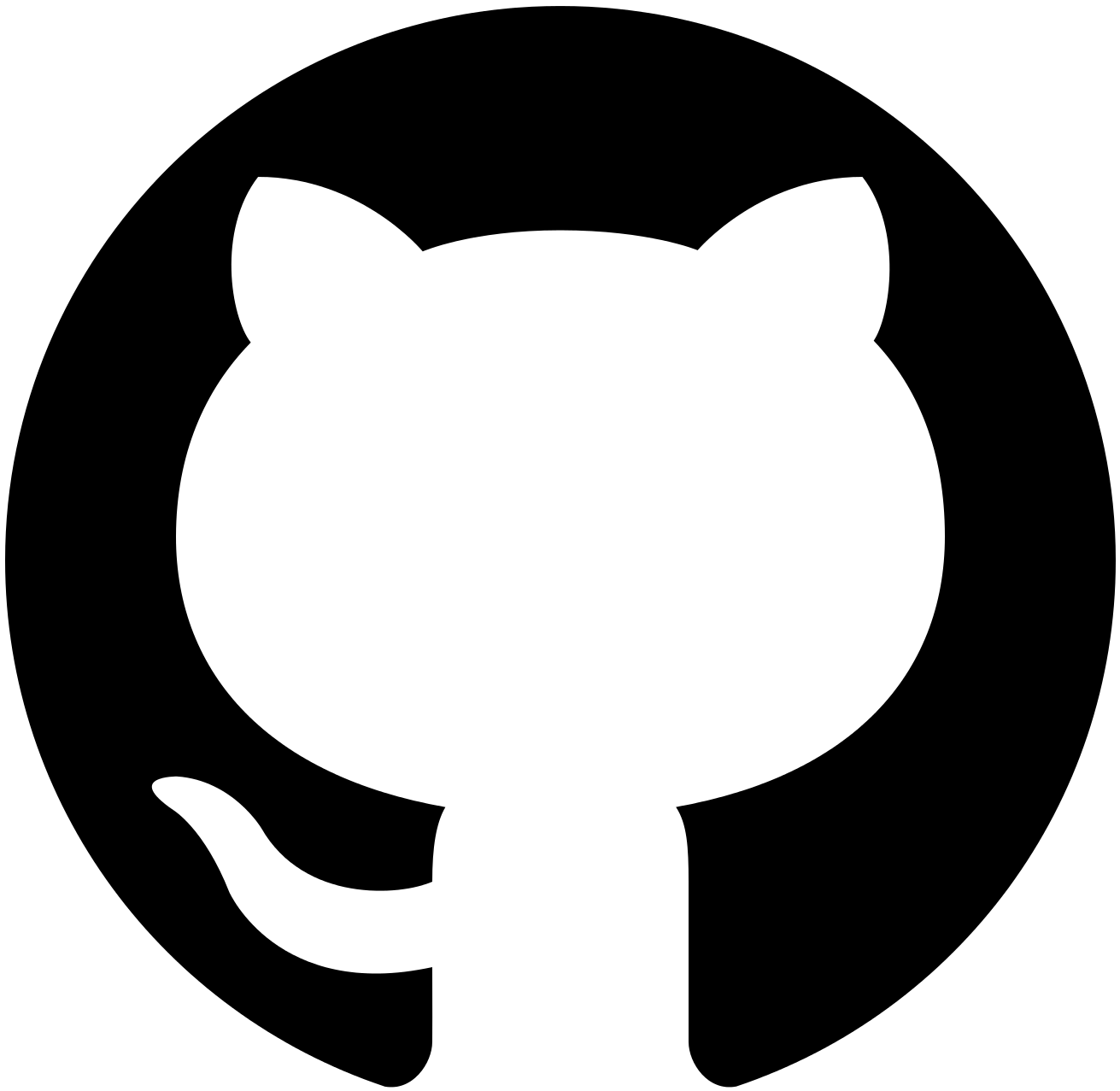}}~~GitHub]{\href{https://github.com/DAGroup-PKU/HumanNet/tree/main/src/model/StableVLA}{\texttt{https://github.com/DAGroup-PKU/HumanNet/tree/main/src/model/StableVLA}}
\\[-1.5ex]}

\checkdata[
\raisebox{-0.2em}{\includegraphics[width=0.025\linewidth]{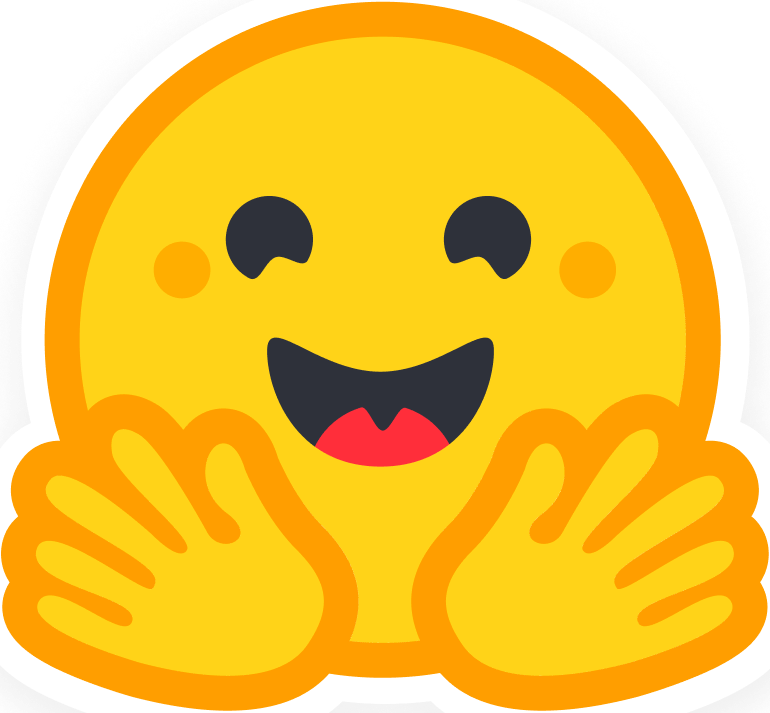}}~~HuggingFace]{\href{https://huggingface.co/DAGroup-PKU/StableVLA}{\texttt{https://huggingface.co/DAGroup-PKU/StableVLA}}
\\[-1.7ex]}

\begin{document}
\maketitle

\section{Introduction}
\label{sec:intro}

The integration of Vision--Language Models (VLMs)~\cite{comanici2025gemini,bai2025qwen2,zhu2025internvl3,xie2024show,li2024llava} into robotic control has fundamentally reshaped the landscape of embodied intelligence. Recent pioneering works~\cite{kim2024openvla,zitkovich2023rt-2,team2024octo,bjorck2025gr00t,black2024pi_0,DBLP:journals/corr/abs-2410-06158} demonstrate that effective alignment among visual perception, large language model (LLM) reasoning, and action execution enables robots to operate across diverse and unstructured scenarios. Building upon this progress, approaches such as VLA-Adapter~\cite{wang2025vla-adapter} propose efficient mechanisms that bridge vision--language representations to the action space through lightweight policy modules, significantly reducing adaptation overhead.

\begin{figure}[htbp]
\centering
\includegraphics[width=\linewidth]{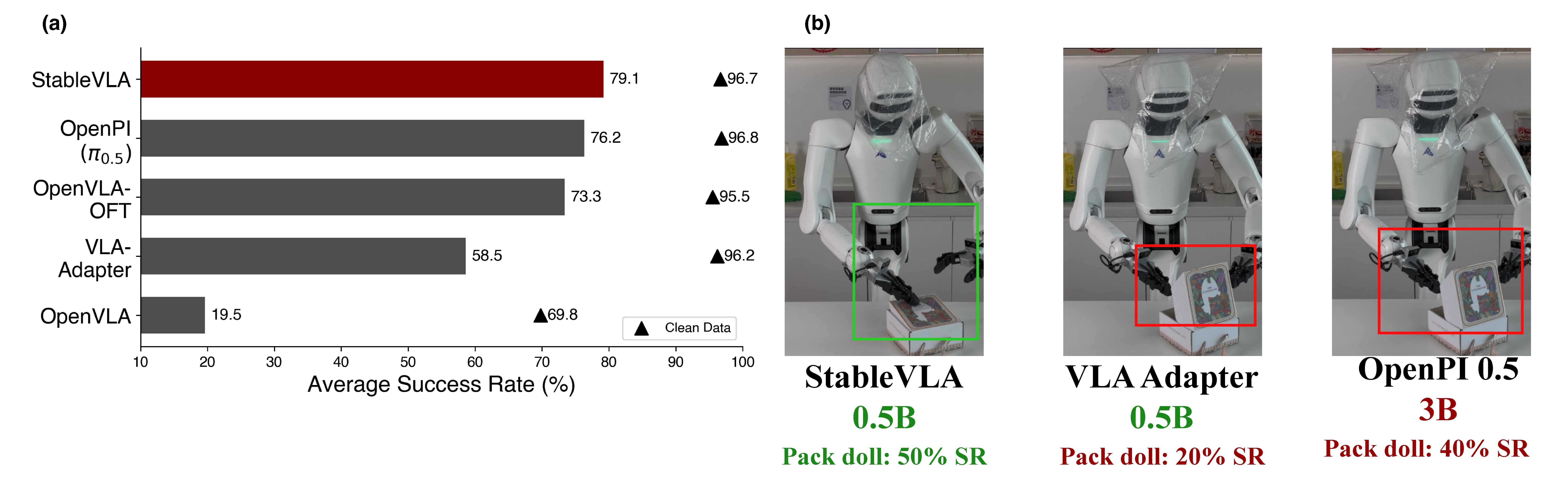}
\caption{\textbf{Overview of StableVLA's robustness.} \textbf{(a)} Robustness comparison on LIBERO benchmark. Triangle marks and bar charts denote performance on clean data and corrupted data (averaged across severities), respectively. StableVLA achieves state-of-the-art zero-shot robustness. \textbf{(b)} Real-world robot deployment on the Pack Doll task under visual corruptions. StableVLA (0.5B) achieves 50\% success rate, outperforming both VLA-Adapter (0.5B, 20\%) and OpenPI~0.5 (3B, 40\%) despite having fewer parameters.}
\label{fig:teaser}
\end{figure}

\begin{wrapfigure}[24]{r}{0.58\linewidth}
\vspace{-1.5em}
\centering
\includegraphics[width=\linewidth]{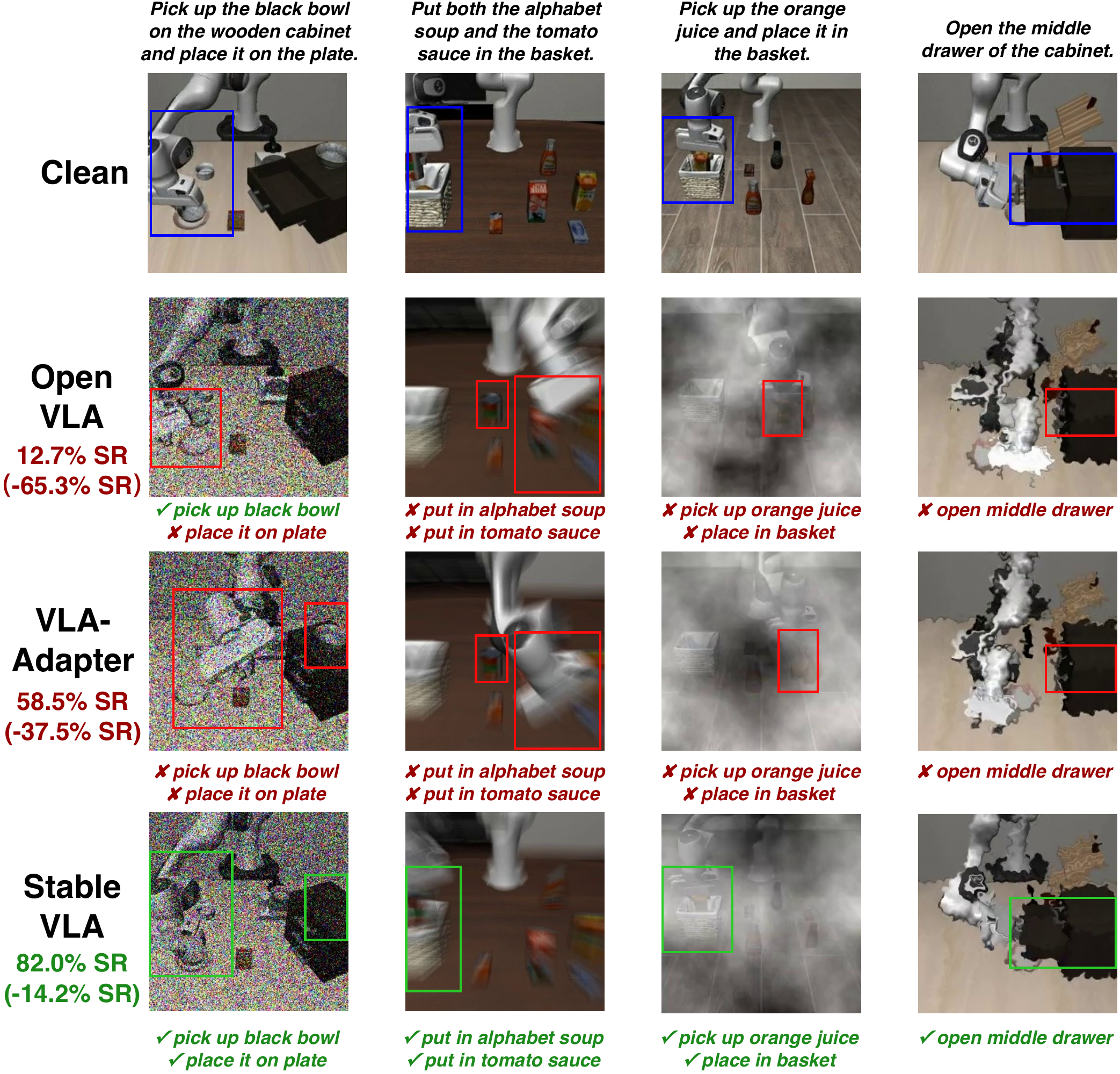}
\caption{\small Catastrophic failure cases under level-5 corruption. StableVLA remains robust while baselines degrade significantly.}
\label{fig:insight}
\vspace{-1.2em}
\end{wrapfigure}

Despite these advances, existing evaluation and benchmarking protocols primarily rely on carefully designed test environments with controlled and idealized visual conditions. In contrast, real-world robotic deployment inevitably involves visual degradations such as sensor noise, motion blur, or weather-induced disturbances, which are largely absent from curated training datasets~\cite{deng2026rethinking,deng2026humannetscalinghumancentricvideo}. This discrepancy introduces a notable gap between model performance observed in benchmark environments and that in real-world settings~\cite{liu2023libero,mees2022calvin,mu2024robotwin}.
Motivated by this gap, we ask the following question: \textit{How do state-of-the-art VLA models perform when exposed to real-world visual disturbances?} To investigate this, we first evaluate the top-performing VLA-Adapter~\cite{wang2025vla-adapter} in simulation by injecting synthetic natural visual corruptions. Surprisingly, a model that originally achieved a high success rate of 96\% experiences nearly a 50\% performance drop under disturbed inputs, as illustrated in Figure~\ref{fig:insight}, and can degrade to 0\% success under certain corruption patterns such as severe visual blur. We further demonstrate that this vulnerability is not unique to VLA-Adapter, but also manifests in other leading VLA models, including OpenVLA~\citep{kim2024openvla}, OpenVLA-OFT~\citep{kim2025fine-tuning}, and OpenPi$\text{--}$0.5~\citep{DBLP:journals/corr/abs-2504-16054}. Consistent performance degradation is also observed in real-world experiments conducted with physical robotic systems, as shown in Figure~\ref{fig:teaser} and Table~\ref{tab:real_res}.

Prevailing strategies for enhancing robustness primarily rely on using extra data with pre-defined distrubations or data augmentation~\citep{hendrycks2021many,DBLP:conf/nips/WangXKYAW21} over clean datasets. However, this data-centric approach faces two fundamental limitations. First, simulating the infinite combinatorial space of real-world corruptions is computationally prohibitive. Second, training with augmented data often induces the memorization of specific noise patterns rather than the learning of robust invariant features, which limits generalization ability to unseen corruptions. This raises a pivotal question: \textit{Can we achieve intrinsic robustness through architectural design, without relying on brute-force data scaling?}

\begin{wrapfigure}[12]{r}{0.43\linewidth}
\vspace{-0.8em}
\centering
\includegraphics[
    width=\linewidth,
    trim=5 10 20 5,
    clip
]{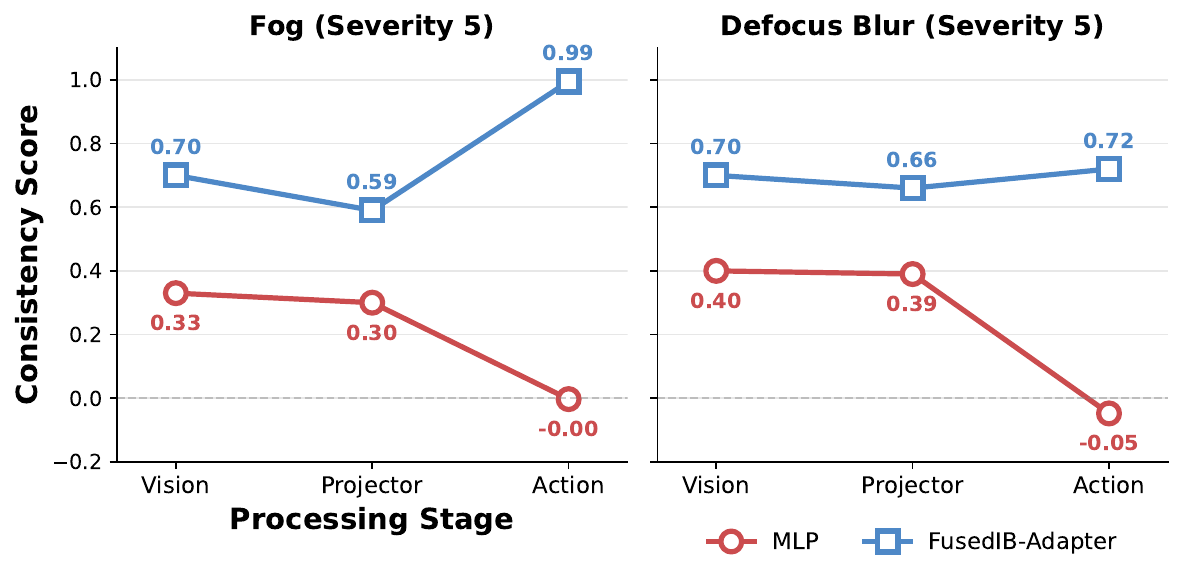}
\caption{\small Feature consistency across VLA processing stages under visual corruptions.}
\label{fig:cascade}
\vspace{-1.0em}
\end{wrapfigure}

We conduct a series of empirical experiments and find evidence suggesting that a significant source of feature vulnerability lies in the projector that bridges the vision encoder and the LLM backbone. As shown in Figure~\ref{fig:cascade}, substantial feature degradation under noisy inputs appears attributable to this projection module.

Motivated by the intrinsic feature selection property of the information bottleneck principle~\citep{tishby2000information}, we propose a novel block structure for connecting the vision branch and the LLM backbone, termed \nameofmethod{}.

By simply replacing the original adapter module in VLA-Adapter and re-training with the same settings, we achieve an average performance improvement of 35.2\% across a range of synthetic visual corruptions. In real-robot experiments, our approach yields a 31.7 percentage point improvement in the pick-and-place task. Owing to its strong robustness against visual disturbances, we refer to the resulting model as \textbf{StableVLA}. Remarkably, StableVLA retains the lightweight training schedule of VLA-Adapter while substantially improving robustness: with only an adapter-level architectural replacement and no additional training data, it surpasses heavily parameterized baselines, including OpenVLA with 14$\times$ more model parameters and $\pi\text{-}0.5$ trained with significantly larger amounts of data.

Our contributions are summarized as follows:
\begin{itemize}[leftmargin=*, topsep=0pt]
\item We conduct empirical studies and observe that current state-of-the-art VLA models, despite achieving strong performance on clean benchmark settings, are highly vulnerable to visual disturbances in both synthetic and real-robot scenarios. Furthermore, our analysis provides evidence that this vulnerability is closely associated with the projection module that bridges the vision encoder and the LLM backbone.
\item We propose a data-free solution by introducing a novel adapter architecture grounded in information bottleneck theory, termed \nameofmethod{}. Under zero-shot settings, simply replacing the original adapter with \nameofmethod{} yields a 35.2\% performance improvement over the baseline in the simulator and 20.4 percentage points on real-robot experiments, while keeping all other experimental settings unchanged.
\item We conduct extensive experiments across multiple benchmarks, including LIBERO~\cite{liu2023libero}, CALVIN~\cite{mees2022calvin}, and real-robot evaluations, on several strong VLA models, such as VLA-Adapter~\cite{wang2025vla-adapter}, OpenVLA~\cite{kim2024openvla}, OpenVLA-OFT~\cite{kim2025fine-tuning}, and $\pi$~\cite{DBLP:journals/corr/abs-2504-16054}. Our results demonstrate that the proposed model consistently outperforms all selected strong baselines while maintaining a significantly smaller model size.
\end{itemize}

\section{Related Work and Preliminaries}
\label{sec:preliminaries}

\subsection{Robustness in Vision Language Models}

Leveraging pre-trained Vision-Language Models (VLMs)~\cite{liu2023visual,comanici2025gemini,liu2024nvila,bai2025qwen2,zhu2025internvl3,xie2024show,li2024llava} for robotic control has become a dominant paradigm in embodied intelligence~\citep{brohan2023rt-1,zitkovich2023rt-2,kim2024openvla,team2024octo,DBLP:conf/rcar/LiWCWSM25,DBLP:journals/ijrr/ChiXFCDBTS25,DBLP:conf/rss/ZhaoKLF23,DBLP:conf/corl/ZawalskiCPMFL24,DBLP:conf/corl/DoshiWMDL24}. Training such models from scratch typically depends on massive datasets, including Open X-Embodiment~\citep{oneill2024open}, DROID~\citep{DBLP:conf/rss/KhazatskyP0BDKN24}, and AgiBot~\cite{contributors2024agibotworldrepo}, and requires substantial computational resources. To alleviate this cost, VLA-Adapter~\citep{wang2025vla-adapter} introduces a resource-efficient architecture that bypasses large-scale pre-training and directly transfers the general perceptual capabilities of VLMs to robotic domains. However, despite improved training efficiency, a critical challenge remains in architectural robustness. In standard VLA models, the vision encoder~\citep{zhai2023sigmoid,oquab2023dinov2} is commonly frozen to preserve semantic priors~\citep{kim2024openvla,kim2025fine-tuning,wang2025vla-adapter}, causing input-level noise or corruption to propagate through the visual backbone. Existing approaches rely on simple MLP-based projectors to align visual features with the policy action space, yet such projectors lack intrinsic mechanisms to suppress task-irrelevant disturbances. Robustness in vision and robotics is traditionally addressed through data-centric strategies, including large-scale data augmentation~\citep{hendrycks2019robustness, hendrycks2021many, DBLP:conf/nips/WangXKYAW21}, and domain randomization in simulation~\citep{tobin2017domain}. However, these methods are computationally expensive and often fail to generalize to unseen perturbations. To overcome these limitations, we propose \textbf{StableVLA}, which targets intrinsic robustness through architectural design by reconstructing the modality alignment interface based on the Information Bottleneck principle, enabling VLA models to effectively filter visual perturbations without relying on exhaustive noise-pattern simulation.

\subsection{Attention Mechanism from the Perspective of Information Bottleneck}

Vision Transformers (ViTs) are more robust to visual corruptions than CNNs~\citep{bai2021transformers,paul2022vision}, a property attributed to self-attention, which promotes \textit{visual grouping} by aggregating tokens into semantic clusters~\citep{zhou2022understanding}. This behavior is theoretically grounded in the Information Bottleneck (IB) principle~\citep{tishby2000information,DBLP:conf/iclr/AlemiFD017}, under which self-attention is shown to be equivalent to iterative IB optimization under Gaussian assumptions~\citep{zhou2022understanding}. Beyond spatial attention, channel-wise grouping has been explored through Cross-Covariance Attention in XCiT~\citep{ali2021xcit} and further interpreted as subspace clustering in FAN~\citep{zhou2022understanding}, where IB-driven channel selection suppresses noise. Building on these insights, \textbf{StableVLA} incorporates a multi-head covariance mechanism into VLA modality alignment to filter noisy channels and enable robust semantic propagation.

\section{Method}
\label{sec:method}

In this section, we present \textbf{StableVLA}, a framework designed to enhance the intrinsic robustness of VLA models. In \cref{sec:preliminaries_ib}, we first formulate the modality alignment problem through the lens of the Information Bottleneck (IB) principle. In \cref{sec:cap}, we introduce the idea of \textbf{Information Bottleneck Adapter (\nameofmethod)}, which utilizes a channel-wise attention mechanism to suppress visual nuisances while preserving task-relevant semantics. In \cref{sec:hybrid}, we further propose our core contribution, \textbf{Fused \nameofmethod}, a hybrid architecture that fuses \nameofmethod~with MLP to retain both robust semantics and fine-grained spatial information critical for precise manipulation.

\begin{figure}[ht]
    \centering
    \includegraphics[width=0.9\linewidth]{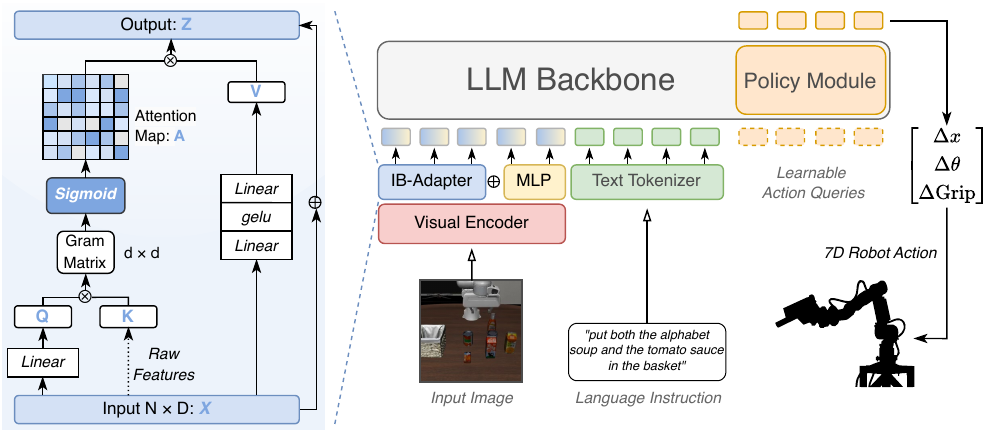}
    \caption{\textbf{Architecture of the \nameofmethod~and StableVLA.} We replace the standard linear bottleneck in StableVLA with Fused {\nameofmethod}. {\nameofmethod} processes visual tokens $\mathbf{X}$ via three parallel pathways to enforce subspace filtering:
    (1) Covariance Attention: We utilize raw features as keys (dotted line) to preserve original geometric structure. We compute the Gram Matrix ($D \times D$) to capture global channel correlations.
    (2) Sigmoid Gating: A learnable Sigmoid function generates the Attention Map $\mathbf{A}$, acting as an independent gate to suppress noise.
    (3) Feature transformation: Features undergo non-linear transformation before reconstruction.}
    \label{fig:cap_arch}
\end{figure}

\subsection{Modality Alignment From an Information Bottleneck Perspective}
\label{sec:preliminaries_ib}

A standard VLA model typically consists of three main parts: a visual encoder $\mathcal{E}$, a learnable projector $\phi$ for modality alignment, and an LLM-based policy model $\pi$. Given a visual observation $\mathbf{I}$ and a text instruction $\mathbf{T}$, the encoder extracts visual tokens $\mathbf{X}_v = \mathcal{E}(\mathbf{I}) \in \mathbb{R}^{N \times D_{v}}$. The projector $\phi$ maps these tokens into the LLM's embedding space: $\mathbf{Z} = \phi(\mathbf{X}_v) \in \mathbb{R}^{N \times D}$. Finally, the LLM predicts actions $\mathbf{a} = \pi(\text{Concat}(\mathbf{Z}, \mathbf{X}_T))$ autoregressively, where $\mathbf{X}_T \in \mathbb{R}^{L \times D}$ represents the text embeddings of $\mathbf{T}$.

In open-world environments, the visual input $\mathbf{I}$ is a composite of task-relevant semantics and task-irrelevant perturbations (e.g., sensor noise). Existing VLA projectors are predominantly implemented as MLP layers. From an Information Bottleneck (IB) perspective, these simple projectors act as \textit{all-pass filters}, which tend to maximize the mutual information $I(\mathbf{X}_v; \mathbf{Z})$ indiscriminately. To enforce intrinsic robustness, we frame modality alignment as an IB problem:
\begin{equation}
    \min_{\phi(\mathbf{Z}|\mathbf{X}_v)} \mathcal{L}_{IB} = I(\mathbf{X}_v; \mathbf{Z}) - \beta I(\mathbf{Z}; \mathbf{S})\label{eq:ib_objective},
\end{equation}
where $\mathbf{Z}$ is a compressed representation that filters nuisances while retaining the target clean code $\mathbf{S}$ (i.e., the ground-truth task-relevant semantics required to predict actions $\mathbf{a}$). The coefficient $\beta$ controls the trade-off between compression and information preservation.

Crucially, while modern ViT-based encoders effectively leverage the IB-driven grouping mechanism in the spatial dimension~\citep{zhou2022understanding}, we argue that for VLA projectors, performing such grouping across the channel dimension ($D$ features) is more critical for robust alignment. Within the visual encoder's output, semantics and noise are often heterogeneously distributed across channels~\citep{zhou2022understanding}. This motivates our  \textbf{\nameofmethod}, which treats each channel as an information unit for IB optimization. By modeling the inter-channel dependencies, {\nameofmethod}  identifies robust semantic subspaces and suppresses uncorrelated noise. Formally,
\begin{proposition}
    \label{prop:ib_sa_channel}
    Let the visual encoder's output $\mathbf{X}_v = [\mathbf{c}_1, \dots, \mathbf{c}_D] \in \mathbb{R}^{N \times D}$ be viewed as a set of $D$ channel-wise observations. Under Gaussian and latent structural assumptions, the iterative update step for the optimal representation $\mathbf{Z}$ that minimizes the IB objective in \cref{eq:ib_objective} corresponds to a channel-wise attention operation:
    \begin{equation}
        \mathbf{Z} = \mathbf{V} \cdot \sigma \left( \beta \mathbf{Q}^\top \mathbf{K} \right),
    \end{equation}
    where $\mathbf{Q, K, V} \in \mathbb{R}^{N \times D}$ are linear projections of $\mathbf{X}_v$. The operator $\sigma(\cdot)$ is a normalizer determined by the latent distribution assumption: $\sigma(\cdot)$ takes the form of \textbf{Softmax} under a categorical latent structure, or \textbf{Sigmoid} under an independent Bernoulli latent structure.
\end{proposition}
The detailed derivation is provided in Appendix \ref{app:proof_ib_sa}.
By extending IB-driven grouping to the channel dimension, this approach enables filtering to dynamically suppress noisy channels while accentuating stable features, establishing representation robustness before features are propagated into the downstream policy model.

\subsection{Information Bottleneck Adapter (\nameofmethod)}
\label{sec:cap}

To enforce the IB principle within the modality alignment stage, we propose the Information Bottleneck Adapter (\nameofmethod). Unlike MLPs that process channels independently, {\nameofmethod} models the inter-channel covariance to identify and amplify robust semantic signals.

Let $\mathbf{X}' \in \mathbb{R}^{N \times D}$ be the input features (e.g., intermediate projector features). The mechanism consists of 3 critical components: subspace covariance modeling, sigmoid-based gating, and non-linear feature transformation.

\paragraph{Subspace covariance modeling.}
We adopt a multi-head design to capture correlations across $H$ diverse semantic subspaces. The input $\mathbf{X}' \in \mathbb{R}^{N \times D}$ is partitioned into $H$ heads $[\mathbf{X}'_1, \dots, \mathbf{X}'_H]$, where each head $\mathbf{X}'_h \in \mathbb{R}^{N \times d}$ has a channel dimension $d = D/H$. For each head $h$, we derive the query $\mathbf{Q}_h = \mathbf{X}'_h \mathbf{W}_q \in \mathbb{R}^{N \times d}$ through a learnable projection $\mathbf{W}_q \in \mathbb{R}^{d \times d}$, while the key $\mathbf{K}_h = \mathbf{X}'_h \in \mathbb{R}^{N \times d}$ is defined via an identity mapping of the input features. This identity-key design ensures that the subsequent covariance computation is grounded in the intrinsic geometric manifold of the visual tokens, thereby preserving high-frequency spatial cues that might otherwise be attenuated by redundant projections. To model inter-channel dependencies, we compute a Gram matrix $\mathbf{G}_h$ by aggregating correlations along the sequence dimension $N$:
\begin{equation}
    \mathbf{G}_h = \mathbf{Q}_h^\top \mathbf{K}_h\in \mathbb{R}^{d \times d} ,
\end{equation}
where each element $\mathbf{G}_h[i, j]$ represents the covariance between channel $i$ and channel $j$ across all spatial tokens.

\paragraph{Sigmoid-based subspace gating.}To separate semantic clusters from independent noise, we apply a learnable sigmoid gating function to the Gram matrix:
\begin{equation}
    \mathbf{A}_h = \sigma\left( \mathbf{G}_h \cdot \boldsymbol{\tau}_h \right) \in [0, 1]^{d \times d},
\end{equation}
where $\boldsymbol{\tau}_h$ is a learnable temperature parameter. The use of sigmoid gating function is theoretically motivated by the \textit{independent Bernoulli latent structure assumption} of the channels. A channel representing uncorrelated sensor noise should exhibit low covariance with semantic-bearing channels, resulting in a gate value near zero. Unlike Softmax operation, which enforces competition between channels by enforcing a categorical distribution over channels, sigmoid gating allows for independent channel selection by suppressing such noisy channels independently without affecting the energy of robust semantic channels.

\textbf{Non-linear feature transformation.}
To enhance feature expressivity, the input $\mathbf{X}_h \in \mathbb{R}^{N \times d}$ is transformed via a two-layer MLP with GELU activation to generate the value tokens $\mathbf{V}_h \in \mathbb{R}^{N \times d}$:
\begin{equation}
    \mathbf{V}_h = \text{Norm}(\text{GELU}(\mathbf{X}_h \mathbf{W}_{v1}) \mathbf{W}_{v2}),
\end{equation}
where $\mathbf{W}_{v1}, \mathbf{W}_{v2} \in \mathbb{R}^{d \times d}$ are learnable weights. The head output $\mathbf{Z}_h \in \mathbb{R}^{N \times d}$ is then reconstructed by modulating these features with the spectral gate $\mathbf{A}_h \in \mathbb{R}^{d \times d}$:
\begin{equation}
    \mathbf{Z}_h = \mathbf{V}_h \mathbf{A}_h,
\end{equation}
and thus $\mathbf{Z}=[\mathbf{Z}_1,\cdots,\mathbf{Z}_H]$. {\nameofmethod} couples non-linear synthesis with channel-wise noise suppression. This design satisfies the IB compression objective (\cref{eq:ib_objective}) by filtering out visual nuisances before the representations are propagated to the LLM backbone.

\subsection{Hybrid Architecture for Balancing Robust Semantics and High-frequency Details}
\label{sec:hybrid}
While {\nameofmethod} effectively suppresses visual disturbance and promotes semantic robustness, it can attenuate high-frequency details essential for precise manipulation. This challenge is particularly evident in long-horizon tasks, where trajectory precision must be maintained over extended sequences. To resolve this trade-off, we propose \textbf{Fused \nameofmethod}, a dual-pathway architecture designed to decouple robust semantic understanding from precise spatial execution:
\begin{equation}
\mathbf{Z} = \text{MLP}(\mathbf{X}) + \tanh(\lambda) \cdot \text{{\nameofmethod}}(\mathbf{X}),
\end{equation}
where $\lambda$ controls the injection of robust signals.
This design maintains two parallel pathways: $i)$ a high-fidelity path using a standard MLP to preserve raw high-frequency details essential for fine-motor control, and $ii)$ a denoising path using the {\nameofmethod} module to extract robust, covariance-filtered semantic features.

\begin{table}[h]
\centering
\setlength{\tabcolsep}{3pt}
\caption{Full comparison on LIBERO and CALVIN benchmark. Methods are grouped by training paradigm. We report success rate (\%). \textbf{Bold}: best, \underline{underline}: second best.}
\label{tab:app_full_libero_calvin}
\resizebox{\textwidth}{!}{%
\begin{tabular}{ll cccc cccc cccc cccc cccc}
\toprule
\multirow{3}{*}{\makecell[l]{Training\\Method}} & \multirow{3}{*}{\makecell[l]{Method}} & \multicolumn{4}{c}{Spatial} & \multicolumn{4}{c}{Object} & \multicolumn{4}{c}{Goal} & \multicolumn{4}{c}{Long} & \multicolumn{4}{c}{CALVIN} \\
\cmidrule(lr){3-6} \cmidrule(lr){7-10} \cmidrule(lr){11-14} \cmidrule(lr){15-18} \cmidrule(lr){19-22}
 & & C & S3 & S4 & S5 & C & S3 & S4 & S5 & C & S3 & S4 & S5 & C & S3 & S4 & S5 & C & S3 & S4 & S5 \\
\midrule
\multirow{2}{*}{\makecell[l]{OpenX\\Pretrain}} & OpenVLA (7B) & 80.0 & 40.9 & 24.6 & 14.7 & 69.6 & 18.2 & 10.4 & 2.7 & 74.0 & 38.7 & 27.0 & 16.3 & 55.5 & 20.5 & 12.4 & 7.0 & -- & -- & -- & -- \\
 & OpenVLA-OFT (7B) & 92.6 & 89.3 & \underline{84.0} & \underline{72.1} & 98.4 & 82.5 & 69.2 & 52.8 & 96.8 & \textbf{94.5} & \underline{84.6} & \underline{70.3} & \textbf{94.4} & \textbf{77.6} & 61.9 & 40.3 & -- & -- & -- & -- \\
\midrule
\makecell[l]{OpenX+Web\\Co-train} & OpenPi--0.5 (3B) & \textbf{98.4} & 88.3 & 79.0 & 62.4 & \textbf{99.4} & \textbf{97.1} & \textbf{88.4} & \textbf{76.4} & 97.2 & 87.2 & 82.5 & 64.2 & 92.0 & 76.1 & \textbf{65.6} & \textbf{47.7} & -- & -- & -- & -- \\
\midrule
\multirow{2}{*}{\makecell[l]{VLM\\Direct FT}} & VLA-Adapter (0.5B) & 96.0 & \underline{93.7} & 83.3 & 58.5 & 96.8 & 71.0 & 44.1 & 29.3 & \underline{97.4} & 79.5 & 64.7 & 47.3 & \underline{94.4} & 63.5 & 41.0 & 26.2 & 4.14 & 2.56 & 1.89 & 1.44 \\
 & StableVLA (0.5B) & \underline{96.2} & \textbf{94.4} & \textbf{92.1} & \textbf{82.0} & \underline{98.8} & \underline{92.4} & \underline{83.6} & \underline{70.2} & \textbf{98.0} & \underline{93.4} & \textbf{85.0} & \textbf{71.9} & 93.6 & \underline{76.3} & \underline{62.4} & \underline{45.3} & \textbf{4.17} & \textbf{2.77} & \textbf{2.11} & \textbf{1.51} \\
\bottomrule
\end{tabular}}
\end{table}

To tailor this balance to specific task dynamics, we calibrate the \textbf{Stochastic Pathway Dropout (SPD)} rate $p_{\text{drop}}$ during fine-tuning.
For tasks demanding extreme spatial fidelity for pick-and-place operations (e.g., LIBERO-Long), retaining the MLP pathway ($p_{\text{drop}} \approx 0$) is crucial. In these scenarios, the IB-Adapter acts as a robustness residual, stabilizing the representation without sacrificing the high-frequency cues required for precise execution.
For tasks requiring consistent object identification or long-horizon semantic planning(e.g., CALVIN, LIBERO-Object), a moderate dropout ($p_{\text{drop}} \approx 0.3$) forces the policy to internalize the robust features from the IB pathway, preventing semantic drift under visual corruptions.
This task-specific configuration allows StableVLA to flexibly navigate the robustness landscape across diverse robotic domains.

\section{Experiments}
\label{sec:experiments}
\subsection{Results on Benchmark}

\subsubsection{Setup}
\paragraph{Benchmarks.}We select the widely adopted LIBERO~\citep{liu2023libero} to evaluate the performance of StableVLA on various types of tasks, and select CALVIN~\citep{mees2022calvin} benchmark to evaluate the zero-shot generalization of StableVLA.
For LIBERO, we utilize all four task categories: LIBERO-Spatial, LIBERO-Object, LIBERO-Goal, and LIBERO-Long.
Each task suit contains 10 subtasks, where each subtask is repeated for 50 episodes for evaluation.
We report the averaged success rate (ranging from 0 to 100\%) over all 500 episodes for each task suit.
For CALVIN, StableVLA is evaluated on environment unseen during training to test its generalization performance.
Specifically, StableVLA is required to execute a predefined sequence of 1,000 tasks in order. Each individual task is composed of five subtasks, and the model may only move on to the subsequent subtask once the current one has been completed.
We report the average completed tasks (ranging from 0 to 5).

\paragraph{Corruption Protocol.}
To rigorously evaluate intrinsic robustness, we adopt the corruption protocol from ImageNet-C~\citep{hendrycks2019robustness}.
We utilize the comprehensive set of corruptions provided by the imagecorruptions library~\citep{michaelis2019dragon}, spanning four categories: noise, blur, weather, and digital corruptions.\footnote{We evaluate the full spectrum of 19 corruptions on LIBERO-Spatial. For LIBERO-Object/Goal/Long and CALVIN benchmarks, we exclude \textit{Glass Blur} due to its prohibitive computational cost during interaction, resulting in 18 corruption types for these tasks.}
These corruptions are defined across 5 severity levels, and we focus our evaluation on the challenging high-severity regime (Levels 3--5) to stress-test architectural stability.
For all evaluations, we conduct experiments over distinct intensity settings (clean, plus severity levels 3, 4, and 5) across the deployed corruption types.

\paragraph{Training Protocol.}
StableVLA replaces the MLP projectors in the VLA-Adapter~\citep{wang2025vla-adapter} framework with our Fused \nameofmethod~module and is trained from scratch on LIBERO and CALVIN.
Detailed hyperparameters, infrastructure specifications, and baseline configurations are provided in Appendix~\ref{app:implementation}.
Following standard VLA training recipes, we apply mild geometric (crop) and photometric (color jitter) augmentations during training to prevent overfitting.
Crucially, we do not expose the models to any aforementioned corruptions or use any specialized robustness techniques (e.g., data augmentation). Thus, the evaluation on the corruptions remains a strictly zero-shot test of architectural generalization.

\paragraph{Baselines.} We benchmark against the state-of-the-art VLA-Adapter~\cite{wang2025vla-adapter}, along with VLAs such as OpenVLA~\citep{kim2024openvla}, OpenVLA-OFT~\citep{kim2025fine-tuning} and OpenPi$\text{--}$0.5~\citep{DBLP:journals/corr/abs-2504-16054}.

\subsubsection{Experiment Results.}
\paragraph{Comprehensive Robustness Profile.} In \cref{tab:app_full_libero_calvin}, we report per-task success rates for LIBERO and average completed tasks for CALVIN, with both metrics averaged over all corruption types.
StableVLA demonstrates superior zero-shot robustness across all benchmarks and corruption severity levels.
On LIBERO, StableVLA consistently scores best or second best across all baselines, rivaling OpenVLA-OFT and OpenPi$\text{--}0.5$ pretrained on large scale datasets like OpenX-Embodiment. It achieves notable improvements over VLA-Adapter that shares similar architecture, especially under severe corruptions, with performance improvements of 40.2\% to 139.6\% across 4 task suites at severity level 5.
On CALVIN, StableVLA consistently completes more tasks than VLA-Adapter across all corruption levels. These results demonstrate the superior zero-shot robustness of StableVLA across various manipulative tasks. More detailed results are provided in Appendix \ref{app:detailed_results}.
\begin{figure}[htbp]
\centering
\begin{subfigure}[c]{0.42\linewidth}
    \centering
    \includegraphics[width=\linewidth]{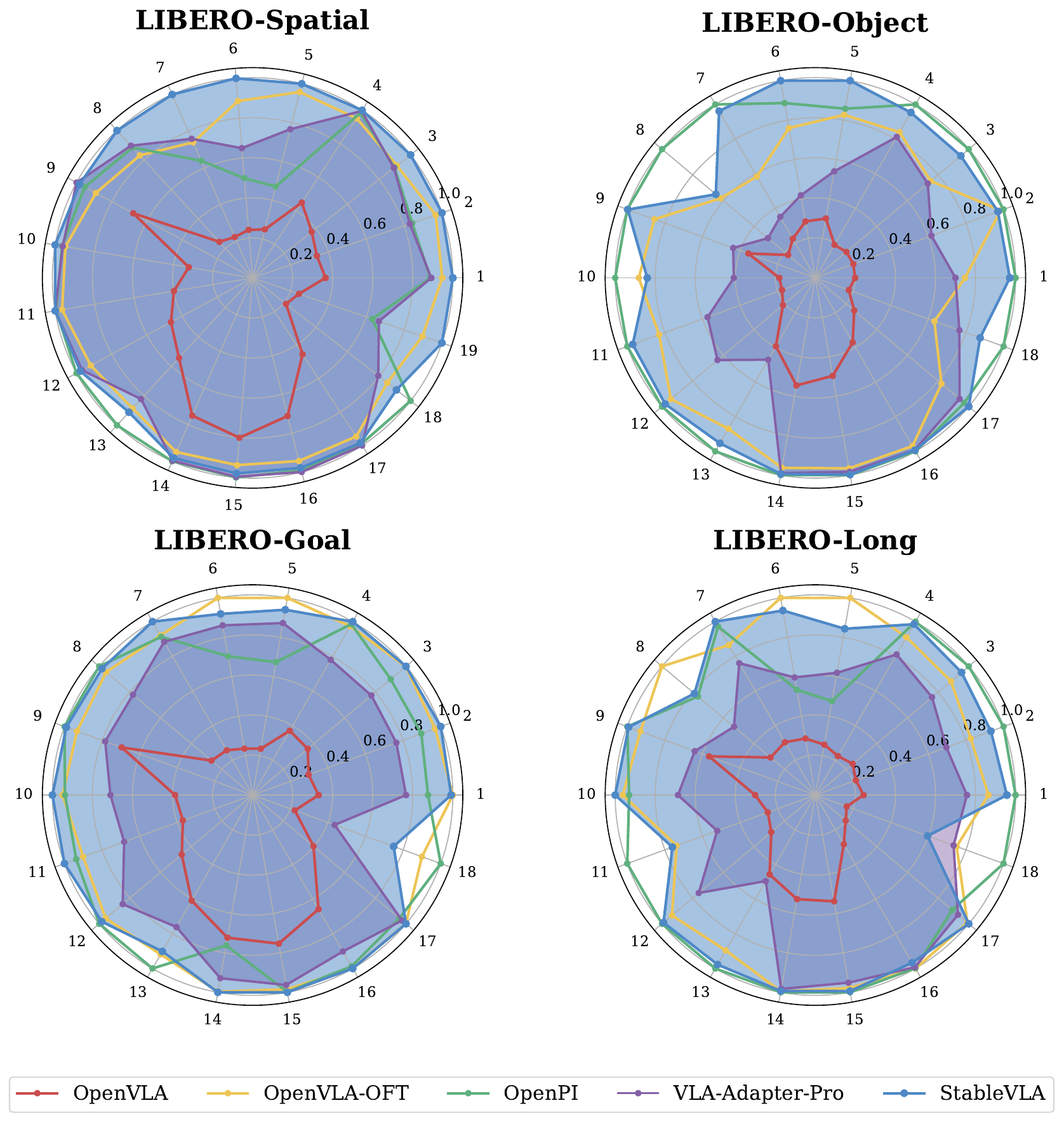}
    \caption{Robustness comparison across corruption types on LIBERO. See Appendix~\ref{app:radar} for details.}
    \label{fig:radar}
\end{subfigure}%
\hfill
\begin{subfigure}[c]{0.56\linewidth}
    \centering
    \includegraphics[width=\linewidth]{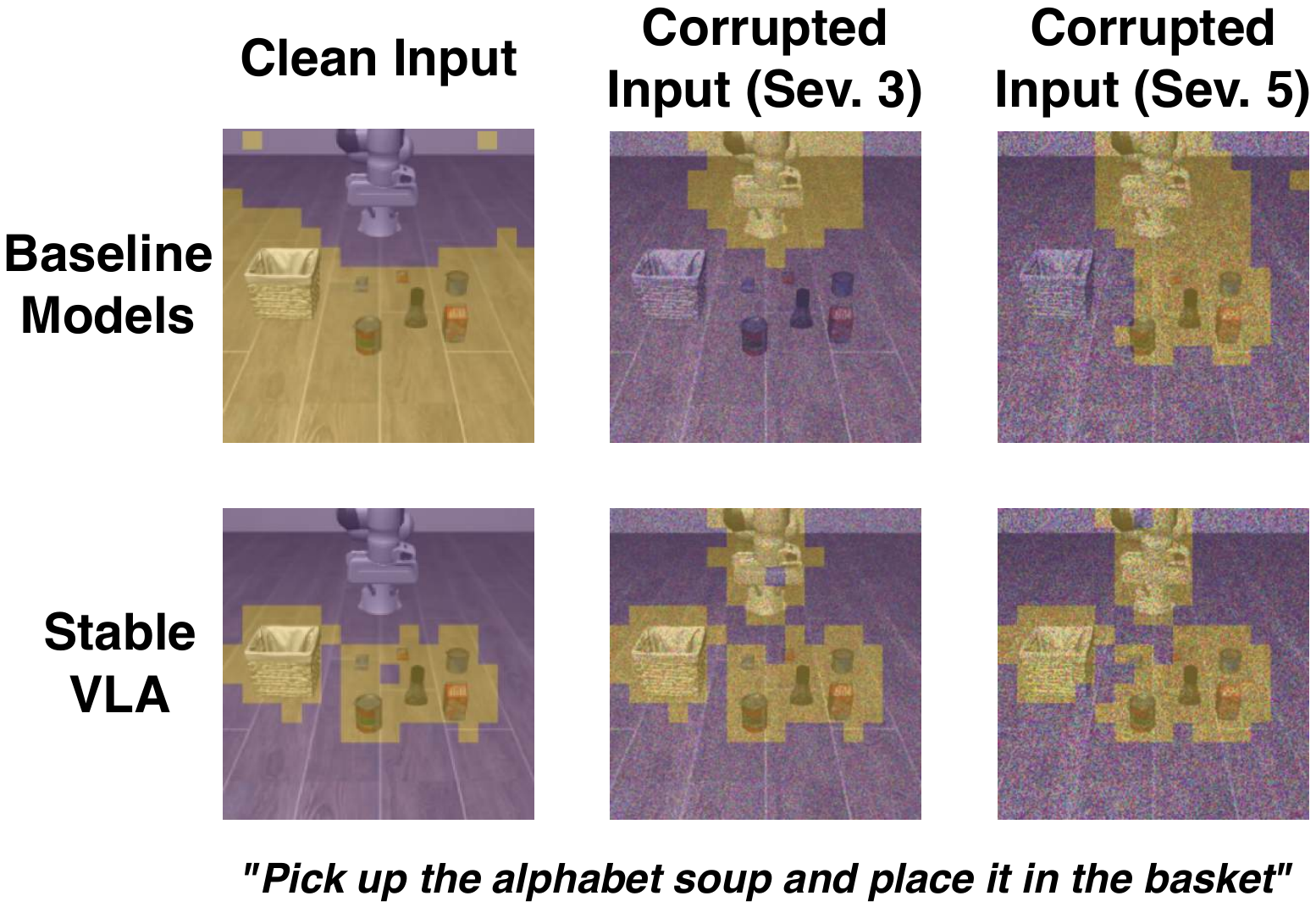}
    \caption{Semantic grouping of Fused \nameofmethod{} features under speckle noise (severity 3 and 5). StableVLA preserves object-centric clusters while baselines degrade.}
    \label{fig:semantic}
\end{subfigure}
\caption{\textbf{Robustness analysis.} (a) Radar charts showing per-corruption robustness across four LIBERO task suites. StableVLA consistently outperforms baselines. (b) K-means clustering ($K\!=\!2$) of output features: standard MLP produces diffused features, while Fused \nameofmethod{} maintains coherent semantic grouping under noise.}
\label{fig:robustness_analysis}
\end{figure}

As illustrated in Figure~\ref{fig:radar}, StableVLA demonstrates a comprehensive robustness advantage, surpassing the VLA-Adapter baseline across the vast majority of corruption categories in all four task suites.
Notably, despite the parameter disparity, StableVLA achieves robustness levels competitive with or surpassing large-scale SOTA models (OpenPi$\text{--}$ and OpenVLA-OFT). This supports the hypothesis that the \nameofmethod~module effectively extracts robust semantic features across diverse manipulation scenarios, helping narrow the performance gap with significantly larger foundation models.

\paragraph{Visualization of Fused IB-Adapter features.}
To demystify the mechanism behind the quantitative gains, we conducted semantic clustering visualization to visualize the features of Fused \nameofmethod~.
Specifically, we conduct K-Means clustering ($K=2$) to the output features of MLP in VLA-Adapter and Fused \nameofmethod~in StableVLA.
For illustration purpose, we examine a manipulation scene from the LIBERO-Object suite under impulse noise, a corruption type characterized by high-frequency, spatially independent disturbances.
Figure~\ref{fig:semantic} demonstrates that even without noise (left column), standard MLPs produce diffused features that conflate task-relevant regions with background. This effect is further amplified under high noise (right columns), which suggests that the standard projector propagates high-frequency disturbances to the downstream policy. In contrast, the output of Fused \nameofmethod~ produces coherent, object-centric semantic groupings. We attribute this to the covariance-based Sigmoid gating in Fused \nameofmethod~. Stochastic corruptions exhibit low correlation with object structures, producing low Gram matrix values that the Sigmoid gates suppress, resulting in a clean focus on grippers and manipulation targets.

\begin{table}[t]
    \begin{minipage}{0.35\linewidth}
        \centering
        \setlength{\lineskip}{1pt} 
        \includegraphics[width=\linewidth]{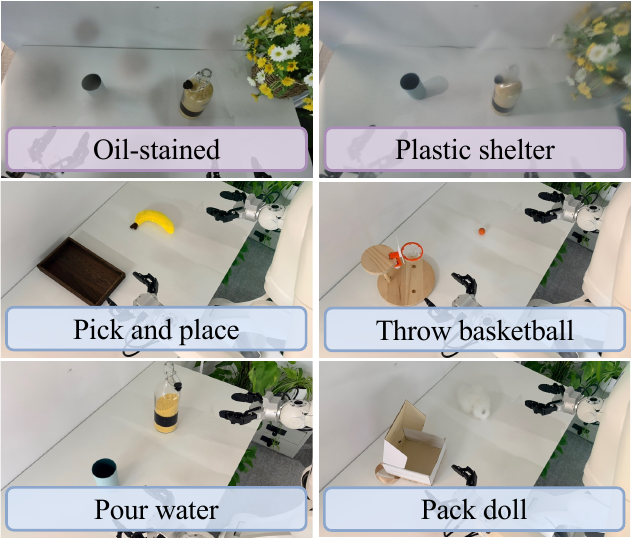}
        \captionof{figure}{\textbf{Visualization of real-world setup.} Top: Physical corruptions (Oil and Shelter). Bottom: Initial states of four evaluation tasks.}
        \label{fig:vis_real}
    \end{minipage}
    \hfill
    % \centering
    \begin{minipage}{0.63\linewidth}
        \centering
        \caption{\textbf{Real-world robustness evaluation.} ``\textbf{Clean}'' reports the absolute success rate (\%). All other columns report the \textbf{performance change} (percentage points, $\Delta$) relative to the clean setting. \textbf{Noise} and \textbf{Blur} represent the average performance drop across different severity levels. \textit{\textbf{Oil}} and \textit{\textbf{Shelter}} denote physical interferences. \textbf{Bold} indicates the best robustness (smallest performance drop).}
        \label{tab:main_results}
        \begin{small}
        \resizebox{\linewidth}{!}{
            \begin{tabular}{ll|c|cccc|c}
            \toprule
            \textbf{Task} & \textbf{Method} & \textbf{Clean} & \textbf{Noise} ($\Delta$) & \textbf{Blur} ($\Delta$) & \textbf{\textit{Oil}} ($\Delta$) & \textbf{\textit{Shelter}} ($\Delta$) & \textbf{Avg.} ($\Delta$) \\
            \midrule

            \multirow{3}{*}{\makecell[l]{Pick\\and place}} 
            & $\pi_{0.5}$        & 100.0 & -63.3 & -16.7 & -10.0 & -30.0 & -30.1 \\
            & VLA-Adapter        & 80.0  & -66.7 & -40.0 & -30.0 & -60.0 & -49.2 \\
            & \textbf{StableVLA} & 80.0  & \textbf{-30.0} & \textbf{-10.0} & \textbf{-10.0} & \textbf{-20.0} & \textbf{-17.5} \\
            \midrule

            \multirow{3}{*}{\makecell[l]{Throw\\basketball}}
            & $\pi_{0.5}$        & 80.0 & -60.0 & -33.3 & -20.0 & -30.0 & -35.8 \\
            & VLA-Adapter        & 60.0 & -53.0 & -40.0 & -20.0 & -40.0 & -38.3 \\
            & \textbf{StableVLA} & 60.0 & \textbf{-36.7} & \textbf{-16.7} & \textbf{-10.0} & \textbf{-10.0} & \textbf{-18.4} \\
            \midrule

            \multirow{3}{*}{Pour water}
            & $\pi_{0.5}$        & 70.0 & -60.0 & -20.0 & -20.0 & -20.0 & -30.0 \\
            & VLA-Adapter        & 40.0 & -40.0 & -30.0 & -10.0 & -20.0 & -25.0 \\
            & \textbf{StableVLA} & 40.0 & \textbf{-23.3} & \textbf{-16.7} & \textbf{-0.0} & \textbf{-10.0} & \textbf{-12.5} \\
            \midrule

            \multirow{3}{*}{Pack doll}
            & $\pi_{0.5}$        & 80.0 & -63.3 & -33.3 & -30.0 & -40.0 & -41.7 \\
            & VLA-Adapter        & 50.0 & -40.0 & -26.7 & -30.0 & -30.0 & -31.7 \\
            & \textbf{StableVLA} & 60.0 & \textbf{-16.7} & \textbf{-10.0} & \textbf{-20.0} & \textbf{-10.0} & \textbf{-14.2} \\

            \bottomrule
            \label{tab:real_res}
            \end{tabular}
        }
        \end{small}
    \end{minipage}
\vspace{-10pt}
\end{table}

\subsection{Real-world Robot Deployment}

\subsubsection{Experiment Setup}
\paragraph{Robot Platform.}
We conduct real-world experiments using the Astribot S1, a high-precision dual-arm robot platform~\cite{gao2025towards}. The robot's chassis and torso are immobilized; control is restricted to the 14-DoF dual arms and gripper actuation. Sensory input consists of RGB streams from a head-mounted camera, which tracks the workspace center, and two wrist-mounted cameras.

\paragraph{Tasks and Data.}
We designed four distinct tasks to evaluate various facets of robotic capability, ranging from basic manipulation to long-horizon planning. A visualization of the task processes is provided in Figure~\ref{fig:real_vis}.
\begin{enumerate}
    \item \textbf{Pick and Place:} Evaluates basic manipulation proficiency. We utilize a set of 5 different objects for comprehensive evaluation, with each object tested over 2 trials. We collected 1000 teleoperated demonstrations for fine-tuning.
    \item \textbf{Throw Basketball:} Evaluates manipulation skills for small objects. We collected 500 teleoperated demonstrations for this task.
    \item \textbf{Pour Water:} Evaluates manipulation precision. We collected 500 teleoperated demonstrations for this task.
    \item \textbf{Pack the Doll:} A long-horizon task requiring multi-stage planning: picking up a doll, placing it precisely into a box, and closing the lid. This tests the model's ability to handle precise geometric constraints. We collected 200 teleoperated demonstrations for fine-tuning.
\end{enumerate}
All expert demonstrations were acquired via VR teleoperation, with the head camera actively tracking the workspace center. Each model is evaluated over 10 trials per task under each corruption setting.

\paragraph{Benchmark Protocol.}
For our real-world experiments, we introduce four types of visual corruptions. First, we simulate two digital corruptions---\textbf{Gaussian noise} and \textbf{defocus blur}---using the \texttt{imagecorruptions} library, applying these effects to the input images prior to model inference. We evaluate Gaussian noise at severity levels 2--4 and defocus blur at levels 3--5, as preliminary experiments indicated that high-severity Gaussian noise leads to catastrophic performance degradation across all baselines. Second, we introduce physical distortions by directly obstructing the camera lens with \textbf{oil} and a \textbf{plastic cover}, separately. The resulting corrupted images are shown in Fig.~\ref{fig:vis_real}. For a fair comparison, we benchmark against $\pi_{0.5}$~\cite{DBLP:journals/corr/abs-2504-16054}, a strong generalist VLA model pretrained on large-scale embodied data, and VLA-Adapter~\cite{wang2025vla-adapter}, which is trained exclusively on our collected dataset.

\subsubsection{Experiment Results}
As shown in Table~\ref{tab:real_res}, while the generalist baseline $\pi_{0.5}$ achieves high success rates on clean data, it suffers significant performance degradation under visual corruptions. VLA-Adapter exhibits similar fragility. In contrast, StableVLA demonstrates superior robustness, consistently maintaining the smallest performance drop across all tasks. Notably, our method shows exceptional resilience against physical interferences (Oil and Shelter), effectively preserving manipulation capabilities where baselines falter.

\newpage

% \input{sections/tables/table_ablation}

% \subsection{Ablation Studies}
% \label{sec:ablation}
% We validate two core design of \nameofmethod~in \cref{tab:ablation}.

% \textbf{Dual-Stream Necessity.}
% As shown in \cref{tab:ablation}, removing the high-fidelity MLP pathway results in performance degradation.
% On LIBERO, the average success rate of \nameofmethod is 3.1\% lower than that of Fused \nameofmethod~on corrupted data.
% On CALVIN, the average number of completed tasks falls from 2.13 to 1.44 when switching from Fused~\nameofmethod to \nameofmethod.
% The dual-stream design proves to be critical: the MLP pathway preserves fine-grained spatial details, while the IB pathway provides semantic robustness, synergizing to achieve optimal performance.

% \textbf{Sigmoid vs. Softmax.}
% As shown in \cref{tab:ablation}, replacing Sigmoid with Softmax leads to a significant performance drop across all benchmarks.
% On LIBERO, the Softmax variant of Fused \nameofmethod witnesses a 16.3-point drop on corrupted data.
% On CALVIN, the average completed tasks collapses from 2.13 to 0.46 on corrupted data.
% This empirically confirms our independent Bernoulli latent structure assumption proposed in \cref{sec:cap}. Unlike Softmax which enforces competition, Sigmoid activation enables independent channel suppression, acting as a more effective spectral filter for uncorrelated noise.
% \label{sec:qualitative}

\begin{wraptable}[10]{l}{0.54\linewidth}
\vspace{-0.6em}
\centering
\footnotesize
\setlength{\tabcolsep}{3pt}
\renewcommand{\arraystretch}{0.92}
\caption{Ablation on adapter architecture.}
\label{tab:ablation}
\begin{tabular*}{\linewidth}{@{\extracolsep{\fill}}llcc@{}}
\toprule
Bench. & Method & Clean & Avg \\
\midrule
\multirow{3}{*}{LIB.}
& IB-Adapter & 96.3 & 76.0 \\
& Fused IB & \textbf{96.6} & \textbf{79.1} \\
& Fused IB-SM & 89.6 & 62.8 \\
\midrule
\multirow{3}{*}{CAL.}
& IB-Adapter & 1.64 & 1.44 \\
& Fused IB & \textbf{4.17} & \textbf{2.13} \\
& Fused IB-SM & 0.46 & 0.46 \\
\bottomrule
\end{tabular*}
\vspace{-0.8em}
\end{wraptable}

\textbf{Dual-Stream Necessity.}
As shown in \cref{tab:ablation}, removing the high-fidelity MLP pathway results in performance degradation.
On LIBERO, the average success rate of \nameofmethod is 3.1 percentage points lower than that of Fused \nameofmethod~on corrupted data.
On CALVIN, the average number of completed tasks falls from 2.13 to 1.44 when switching from Fused~\nameofmethod to \nameofmethod.
The dual-stream design proves critical: the MLP pathway preserves fine-grained spatial details, while the IB pathway provides semantic robustness.

\textbf{Sigmoid vs. Softmax.}
As shown in \cref{tab:ablation}, replacing Sigmoid with Softmax leads to a significant performance drop across all benchmarks.
On LIBERO, the Softmax variant of Fused \nameofmethod drops by 16.3 percentage points on corrupted data.
On CALVIN, the average completed tasks collapse from 2.13 to 0.46.
This empirically confirms our independent Bernoulli latent structure assumption proposed in \cref{sec:cap}.
Unlike Softmax, which enforces competition, Sigmoid activation enables independent channel suppression, acting as a more effective spectral filter for uncorrelated noise.
Here, SM denotes Softmax.

\section{Conclusions}
\label{sec:conclusion}

In this work, we addressed a notable vulnerability of VLA modality alignment to visual corruptions. Guided by the Information Bottleneck principle, we proposed the \textbf{\nameofmethod}, a general-purpose dual-stream architecture that synergizes high-fidelity spatial processing with covariance-based spectral filtering. Extensive evaluations confirm that this design confers superior zero-shot robustness across diverse benchmarks. Remarkably, our results demonstrate that \textbf{architectural innovation can help narrow the scaling gap}: despite using a 14$\times$ smaller backbone without external OpenX pre-training, our approach achieves robustness competitive with data-intensive 7B-scale SOTA models. As a parameter-efficient and potentially \textbf{model-agnostic} component, the \nameofmethod~offers a compelling alternative to standard projectors, encouraging a shift towards robust architectural inductive biases in the development of next-generation embodied AI.

\section*{Acknowledgements}
We gratefully acknowledge Simple Silicon Innovation for their valuable support throughout this project. 
We also thank the National Natural Science Foundation of China (NSFC) for partially supporting Qibin Hou under Grant No.~62522607.
\clearpage

\bibliographystyle{plainnat}
\setlength{\bibhang}{0pt}
\setlength\bibindent{0pt}
\bibliography{main}

@inproceedings{brohan2023rt-1,
	title        = {RT-1: Robotics Transformer for Real-World Control at Scale.},
	author       = {Brohan, Anthony and Brown, Noah and Carbajal, Justice and Chebotar, Yevgen and Dabis, Joseph and Finn, Chelsea and Gopalakrishnan, Keerthana and Hausman, Karol and others},
	booktitle    = {Robotics: Science and Systems},
	year         = 2023,
}

@inproceedings{oneill2024open,
	title        = {Open X-Embodiment: Robotic Learning Datasets and RT-X Models : Open X-Embodiment Collaboration.},
	author       = {O'Neill, Abby and Rehman, Abdul and Maddukuri, Abhiram and Gupta, Abhishek and Padalkar, Abhishek and Lee, Abraham and Pooley, Acorn and Gupta, Agrim and Mandlekar, Ajay and others},
	booktitle    = {ICRA},
	year         = 2024,
}

@article{wang2025vla-adapter,
	title        = {VLA-Adapter: An Effective Paradigm for Tiny-Scale Vision-Language-Action Model.},
	author       = {Wang, Yihao and Ding, Pengxiang and Li, Lingxiao and Cui, Can and Ge, Zirui and Tong, Xinyang and Song, Wenxuan and Zhao, Han and Zhao, Wei and Hou, Pengxu and Huang, Siteng and Tang, Yifan and Wang, Wenhui and Zhang, Ru and Liu, Jianyi and Wang, Donglin},
	year         = 2025,
	journal      = {CoRR},
}

@article{kim2025fine-tuning,
	title        = {Fine-Tuning Vision-Language-Action Models: Optimizing Speed and Success.},
	author       = {Kim, Moo Jin and Finn, Chelsea and Liang, Percy},
	year         = 2025,
	journal      = {CoRR},
}

@inproceedings{zhou2022understanding,
  title={Understanding the robustness in vision transformers},
  author={Zhou, Daquan and Yu, Zhiding and Xie, Enze and Xiao, Chaowei and Anandkumar, Animashree and Feng, Jiashi and Alvarez, Jose M},
  booktitle={International conference on machine learning},
  pages={27378--27394},
  year={2022},
  organization={PMLR}
}

@article{ali2021xcit,
  title={Xcit: Cross-covariance image transformers},
  author={Ali, Alaaeldin and Touvron, Hugo and Caron, Mathilde and Bojanowski, Piotr and Douze, Matthijs and Joulin, Armand and Laptev, Ivan and Neverova, Natalia and Synnaeve, Gabriel and Verbeek, Jakob and others},
  journal={Advances in neural information processing systems},
  volume={34},
  pages={20014--20027},
  year={2021}
}

@article{tishby2000information,
  title={The information bottleneck method},
  author={Tishby, Naftali and Pereira, Fernando C and Bialek, William},
  journal={arXiv preprint physics/0004057},
  year={2000}
}

@article{team2024octo,
  title={Octo: An open-source generalist robot policy},
  author={Team, Octo Model and Ghosh, Dibya and Walke, Homer and Pertsch, Karl and Black, Kevin and Mees, Oier and Dasari, Sudeep and Hejna, Joey and Kreiman, Tobias and Xu, Charles and others},
  journal={arXiv preprint arXiv:2405.12213},
  year={2024}
}

@inproceedings{zhai2023sigmoid,
  title={Sigmoid loss for language image pre-training},
  author={Zhai, Xiaohua and Mustafa, Basil and Kolesnikov, Alexander and Beyer, Lucas},
  booktitle={Proceedings of the IEEE/CVF international conference on computer vision},
  pages={11975--11986},
  year={2023}
}

@article{oquab2023dinov2,
  title={Dinov2: Learning robust visual features without supervision},
  author={Oquab, Maxime and Darcet, Timoth{\'e}e and Moutakanni, Th{\'e}o and Vo, Huy and Szafraniec, Marc and Khalidov, Vasil and Fernandez, Pierre and Haziza, Daniel and Massa, Francisco and El-Nouby, Alaaeldin and others},
  journal={arXiv preprint arXiv:2304.07193},
  year={2023}
}

@article{hendrycks2019robustness,
  title={Benchmarking Neural Network Robustness to Common Corruptions and Perturbations},
  author={Dan Hendrycks and Thomas Dietterich},
  journal={Proceedings of the International Conference on Learning Representations},
  year={2019}
}

@article{hendrycks2021many,
  title={The Many Faces of Robustness: A Critical Analysis of Out-of-Distribution Generalization},
  author={Dan Hendrycks and Steven Basart and Norman Mu and Saurav Kadavath and Frank Wang and Evan Dorundo and Rahul Desai and Tyler Zhu and Samyak Parajuli and Mike Guo and Dawn Song and Jacob Steinhardt and Justin Gilmer},
  journal={ICCV},
  year={2021}
}

@inproceedings{tobin2017domain,
  title={Domain randomization for transferring deep neural networks from simulation to the real world},
  author={Tobin, Josh and Fong, Rachel and Ray, Alex and Schneider, Jonas and Zaremba, Wojciech and Abbeel, Pieter},
  booktitle={2017 IEEE/RSJ international conference on intelligent robots and systems (IROS)},
  pages={23--30},
  year={2017},
  organization={IEEE}
}

@article{bai2021transformers,
  title={Are transformers more robust than cnns?},
  author={Bai, Yutong and Mei, Jieru and Yuille, Alan L and Xie, Cihang},
  journal={Advances in neural information processing systems},
  volume={34},
  pages={26831--26843},
  year={2021}
}

@inproceedings{paul2022vision,
  title={Vision transformers are robust learners},
  author={Paul, Sayak and Chen, Pin-Yu},
  booktitle={Proceedings of the AAAI conference on Artificial Intelligence},
  volume={36},
  number={2},
  pages={2071--2081},
  year={2022}
}

@article{mees2022calvin,
author = {Oier Mees and Lukas Hermann and Erick Rosete-Beas and Wolfram Burgard},
title = {CALVIN: A Benchmark for Language-Conditioned Policy Learning for Long-Horizon Robot Manipulation Tasks},
journal={IEEE Robotics and Automation Letters (RA-L)},
volume={7},
number={3},
pages={7327-7334},
year={2022}
}

@inproceedings{mu2024robotwin,
  title={Robotwin: Dual-arm robot benchmark with generative digital twins (early version)},
  author={Mu, Yao and Chen, Tianxing and Peng, Shijia and Chen, Zanxin and Gao, Zeyu and Zou, Yude and Lin, Lunkai and Xie, Zhiqiang and Luo, Ping},
  booktitle={European Conference on Computer Vision},
  pages={264--273},
  year={2024},
  organization={Springer}
}

@inproceedings{karamcheti2024prismatic,
  title = {Prismatic VLMs: Investigating the Design Space of Visually-Conditioned Language Models},
  author = {Siddharth Karamcheti and Suraj Nair and Ashwin Balakrishna and Percy Liang and Thomas Kollar and Dorsa Sadigh},
  booktitle = {International Conference on Machine Learning (ICML)},
  year = {2024},
}

@article{liu2023libero,
  title={Libero: Benchmarking knowledge transfer for lifelong robot learning},
  author={Liu, Bo and Zhu, Yifeng and Gao, Chongkai and Feng, Yihao and Liu, Qiang and Zhu, Yuke and Stone, Peter},
  journal={Advances in Neural Information Processing Systems},
  volume={36},
  pages={44776--44791},
  year={2023}
}

@article{michaelis2019dragon,
  title={Benchmarking Robustness in Object Detection: 
    Autonomous Driving when Winter is Coming},
  author={Michaelis, Claudio and Mitzkus, Benjamin and 
    Geirhos, Robert and Rusak, Evgenia and 
    Bringmann, Oliver and Ecker, Alexander S. and 
    Bethge, Matthias and Brendel, Wieland},
  journal={arXiv preprint arXiv:1907.07484},
  year={2019}
}

@article{DBLP:journals/corr/abs-2504-16054,
  author       = {Physical Intelligence and
                  Kevin Black and
                  Noah Brown and
                  James Darpinian and
                  Karan Dhabalia and
                  Danny Driess and
                  Adnan Esmail and
                  Michael Equi and
                  Chelsea Finn and
                  Niccolo Fusai and
                  Manuel Y. Galliker and
                  Dibya Ghosh and
                  Lachy Groom and
                  Karol Hausman and
                  Brian Ichter and
                  Szymon Jakubczak and
                  Tim Jones and
                  Liyiming Ke and
                  Devin LeBlanc and
                  Sergey Levine and
                  Adrian Li{-}Bell and
                  Mohith Mothukuri and
                  Suraj Nair and
                  Karl Pertsch and
                  Allen Z. Ren and
                  Lucy Xiaoyang Shi and
                  Laura Smith and
                  Jost Tobias Springenberg and
                  Kyle Stachowicz and
                  James Tanner and
                  Quan Vuong and
                  Homer Walke and
                  Anna Walling and
                  Haohuan Wang and
                  Lili Yu and
                  Ury Zhilinsky},
  title        = {{\(\pi\)}\({}_{\mbox{0.5}}\): a Vision-Language-Action Model with
                  Open-World Generalization},
  journal      = {CoRR},
year         = {2025}
}

@article{gao2025towards,
  title={Towards Human-level Intelligence via Human-like Whole-Body Manipulation},
  author={Gao, Guang and Wang, Jianan and Zuo, Jinbo and Jiang, Junnan and Zhang, Jingfan and Zeng, Xianwen and Zhu, Yuejiang and Ma, Lianyang and Chen, Ke and Sheng, Minhua and others},
  journal={arXiv preprint arXiv:2507.17141},
  year={2025}
}

@misc{contributors2024agibotworldrepo,
  title={AgiBot World Colosseum},
  author={AgiBot World Colosseum contributors},
  howpublished={\url{https://github.com/OpenDriveLab/AgiBot-World}},
  year={2024}
}

@article{liu2023visual,
  title={Visual instruction tuning},
  author={Liu, Haotian and Li, Chunyuan and Wu, Qingyang and Lee, Yong Jae},
  journal={Advances in neural information processing systems},
  volume={36},
  pages={34892--34916},
  year={2023}
}

@article{comanici2025gemini,
  title={Gemini 2.5: Pushing the frontier with advanced reasoning, multimodality, long context, and next generation agentic capabilities},
  author={Comanici, Gheorghe and Bieber, Eric and Schaekermann, Mike and Pasupat, Ice and Sachdeva, Noveen and Dhillon, Inderjit and Blistein, Marcel and Ram, Ori and Zhang, Dan and Rosen, Evan and others},
  journal={arXiv preprint arXiv:2507.06261},
  year={2025}
}

@article{liu2024nvila,
  title={NVILA: Efficient frontier visual language models},
  author={Liu, Zhijian and Zhu, Ligeng and Shi, Baifeng and Zhang, Zhuoyang and Lou, Yuming and Yang, Shang and Xi, Haocheng and Cao, Shiyi and Gu, Yuxian and Li, Dacheng and others},
  journal={arXiv preprint arXiv:2412.04468},
  year={2024}
}

@article{bai2025qwen2,
  title={Qwen2. 5-vl technical report},
  author={Bai, Shuai and Chen, Keqin and Liu, Xuejing and Wang, Jialin and Ge, Wenbin and Song, Sibo and Dang, Kai and Wang, Peng and Wang, Shijie and Tang, Jun and others},
  journal={arXiv preprint arXiv:2502.13923},
  year={2025}
}

@article{xie2024show,
  title={Show-o: One single transformer to unify multimodal understanding and generation},
  author={Xie, Jinheng and Mao, Weijia and Bai, Zechen and Zhang, David Junhao and Wang, Weihao and Lin, Kevin Qinghong and Gu, Yuchao and Chen, Zhijie and Yang, Zhenheng and Shou, Mike Zheng},
  journal={arXiv preprint arXiv:2408.12528},
  year={2024}
}

@article{deng2026rethinking,
  title={Rethinking Video Generation Model for the Embodied World},
  author={Deng, Yufan and Pan, Zilin and Zhang, Hongyu and Li, Xiaojie and Hu, Ruoqing and Ding, Yufei and Zou, Yiming and Zeng, Yan and Zhou, Daquan},
  journal={arXiv preprint arXiv:2601.15282},
  year={2026}
}

@misc{deng2026humannetscalinghumancentricvideo,
      title={HumanNet: Scaling Human-centric Video Learning to One Million Hours}, 
      author={Yufan Deng and Daquan Zhou},
      year={2026},
      eprint={2605.06747},
      archivePrefix={arXiv},
      primaryClass={cs.CV},
      url={https://arxiv.org/abs/2605.06747}, 
}

@article{zhu2025internvl3,
  title={Internvl3: Exploring advanced training and test-time recipes for open-source multimodal models},
  author={Zhu, Jinguo and Wang, Weiyun and Chen, Zhe and Liu, Zhaoyang and Ye, Shenglong and Gu, Lixin and Tian, Hao and Duan, Yuchen and Su, Weijie and Shao, Jie and others},
  journal={arXiv preprint arXiv:2504.10479},
  year={2025}
}

@article{li2024llava,
  title={Llava-onevision: Easy visual task transfer},
  author={Li, Bo and Zhang, Yuanhan and Guo, Dong and Zhang, Renrui and Li, Feng and Zhang, Hao and Zhang, Kaichen and Zhang, Peiyuan and Li, Yanwei and Liu, Ziwei and others},
  journal={arXiv preprint arXiv:2408.03326},
  year={2024}
}

@inproceedings{kim2024openvla,
	title        = {OpenVLA: An Open-Source Vision-Language-Action Model.},
	author       = {Kim, Moo Jin and Pertsch, Karl and Karamcheti, Siddharth and Xiao, Ted and Balakrishna, Ashwin and Nair, Suraj and Rafailov, Rafael and Foster, Ethan Paul and Sanketi, Pannag R. and Vuong, Quan and Kollar, Thomas and Burchfiel, Benjamin and Tedrake, Russ and Sadigh, Dorsa and Levine, Sergey and Liang, Percy and Finn, Chelsea},
	booktitle    = {CoRL},
	year         = 2024,
}

@inproceedings{zitkovich2023rt-2,
	title        = {RT-2: Vision-Language-Action Models Transfer Web Knowledge to Robotic Control.},
	author       = {Zitkovich, Brianna and Yu, Tianhe and Xu, Sichun and Xu, Peng and Xiao, Ted and Xia, Fei and Wu, Jialin and Wohlhart, Paul and Welker, Stefan and Wahid, Ayzaan and others},
	booktitle    = {CoRL},
	year         = 2023,
}

@article{bjorck2025gr00t,
  title={Gr00t n1: An open foundation model for generalist humanoid robots},
  author={Bjorck, Johan and Casta{\~n}eda, Fernando and Cherniadev, Nikita and Da, Xingye and Ding, Runyu and Fan, Linxi and Fang, Yu and Fox, Dieter and Hu, Fengyuan and Huang, Spencer and others},
  journal={arXiv preprint arXiv:2503.14734},
  year={2025}
}

@article{black2024pi_0,
  title={$\pi_0$: A Vision-Language-Action Flow Model for General Robot Control},
  author={Black, Kevin and Brown, Noah and Driess, Danny and Esmail, Adnan and Equi, Michael and Finn, Chelsea and Fusai, Niccolo and Groom, Lachy and Hausman, Karol and Ichter, Brian and others},
  journal={arXiv preprint arXiv:2410.24164},
  year={2024}
}

@inproceedings{DBLP:conf/nips/WangXKYAW21,
  author       = {Haotao Wang and
                  Chaowei Xiao and
                  Jean Kossaifi and
                  Zhiding Yu and
                  Anima Anandkumar and
                  Zhangyang Wang},
  editor       = {Marc'Aurelio Ranzato and
                  Alina Beygelzimer and
                  Yann N. Dauphin and
                  Percy Liang and
                  Jennifer Wortman Vaughan},
  title        = {AugMax: Adversarial Composition of Random Augmentations for Robust
                  Training},
  booktitle    = {Advances in Neural Information Processing Systems 34: Annual Conference
                  on Neural Information Processing Systems 2021, NeurIPS 2021, December
                  6-14, 2021, virtual},
  pages        = {237--250},
  year         = {2021},
  url          = {https://proceedings.neurips.cc/paper/2021/hash/01e9565cecc4e989123f9620c1d09c09-Abstract.html},
  bibsource    = {dblp computer science bibliography, https://dblp.org}
}

@inproceedings{DBLP:conf/iclr/AlemiFD017,
  author       = {Alexander A. Alemi and
                  Ian Fischer and
                  Joshua V. Dillon and
                  Kevin Murphy},
  title        = {Deep Variational Information Bottleneck},
  booktitle    = {5th International Conference on Learning Representations, {ICLR} 2017,
                  Toulon, France, April 24-26, 2017, Conference Track Proceedings},
  publisher    = {OpenReview.net},
  year         = {2017},
  url          = {https://openreview.net/forum?id=HyxQzBceg},
  bibsource    = {dblp computer science bibliography, https://dblp.org}
}

@article{DBLP:journals/corr/abs-2410-06158,
  author       = {Chilam Cheang and
                  Guangzeng Chen and
                  Ya Jing and
                  Tao Kong and
                  Hang Li and
                  Yifeng Li and
                  Yuxiao Liu and
                  Hongtao Wu and
                  Jiafeng Xu and
                  Yichu Yang and
                  Hanbo Zhang and
                  Minzhao Zhu},
  title        = {{GR-2:} {A} Generative Video-Language-Action Model with Web-Scale
                  Knowledge for Robot Manipulation},
  journal      = {CoRR},
  volume       = {abs/2410.06158},
  year         = {2024},
  url          = {https://doi.org/10.48550/arXiv.2410.06158},
  doi          = {10.48550/ARXIV.2410.06158},
  eprinttype   = {arXiv},
  eprint       = {2410.06158},
  bibsource    = {dblp computer science bibliography, https://dblp.org}
}

@inproceedings{DBLP:conf/rcar/LiWCWSM25,
  author       = {Xiaojian Li and
                  Sheng Wang and
                  Chao Chen and
                  Hailong Wei and
                  Yudong Shi and
                  Hangjie Mo},
  title        = {RoboFlamingo-Plus: Fusion of Depth and {RGB} Perception with Vision-Language
                  Models for Enhanced Robotic Manipulation},
  booktitle    = {{IEEE} International Conference on Real-time Computing and Robotics,
                  {RCAR} 2025, Toyama, Japan, June 1-6, 2025},
  pages        = {311--316},
  publisher    = {{IEEE}},
  year         = {2025},
  url          = {https://doi.org/10.1109/RCAR65431.2025.11139480},
  doi          = {10.1109/RCAR65431.2025.11139480},
  bibsource    = {dblp computer science bibliography, https://dblp.org}
}

@inproceedings{DBLP:conf/rss/KhazatskyP0BDKN24,
  author       = {Alexander Khazatsky and
                  Karl Pertsch and
                  Suraj Nair and
                  Ashwin Balakrishna and
                  Sudeep Dasari and
                  Siddharth Karamcheti and
                  Soroush Nasiriany and
                  Mohan Kumar Srirama and
                  Lawrence Yunliang Chen and
                  Kirsty Ellis and
                  Peter David Fagan and
                  Joey Hejna and
                  Masha Itkina and
                  Marion Lepert and
                  Yecheng Jason Ma and
                  Patrick Tree Miller and
                  Jimmy Wu and
                  Suneel Belkhale and
                  Shivin Dass and
                  Huy Ha and
                  Arhan Jain and
                  Abraham Lee and
                  Youngwoon Lee and
                  Marius Memmel and
                  Sungjae Park and
                  Ilija Radosavovic and
                  Kaiyuan Wang and
                  Albert Zhan and
                  Kevin Black and
                  Cheng Chi and
                  Kyle Beltran Hatch and
                  Shan Lin and
                  Jingpei Lu and
                  Jean Mercat and
                  Abdul Rehman and
                  Pannag R. Sanketi and
                  Archit Sharma and
                  Cody Simpson and
                  Quan Vuong and
                  Homer Rich Walke and
                  Blake Wulfe and
                  Ted Xiao and
                  Jonathan Heewon Yang and
                  Arefeh Yavary and
                  Tony Z. Zhao and
                  Christopher Agia and
                  Rohan Baijal and
                  Mateo Guaman Castro and
                  Daphne Chen and
                  Qiuyu Chen and
                  Trinity Chung and
                  Jaimyn Drake and
                  Ethan Paul Foster and
                  Jensen Gao and
                  David Antonio Herrera and
                  Minho Heo and
                  Kyle Hsu and
                  Jiaheng Hu and
                  Donovon Jackson and
                  Charlotte Le and
                  Yunshuang Li and
                  Roy Lin and
                  Zehan Ma and
                  Abhiram Maddukuri and
                  Suvir Mirchandani and
                  Daniel Morton and
                  Tony Nguyen and
                  Abigail O'Neill and
                  Rosario Scalise and
                  Derick Seale and
                  Victor Son and
                  Stephen Tian and
                  Emi Tran and
                  Andrew E. Wang and
                  Yilin Wu and
                  Annie Xie and
                  Jingyun Yang and
                  Patrick Yin and
                  Yunchu Zhang and
                  Osbert Bastani and
                  Glen Berseth and
                  Jeannette Bohg and
                  Ken Goldberg and
                  Abhinav Gupta and
                  Abhishek Gupta and
                  Dinesh Jayaraman and
                  Joseph J. Lim and
                  Jitendra Malik and
                  Roberto Mart{\'{\i}}n{-}Mart{\'{\i}}n and
                  Subramanian Ramamoorthy and
                  Dorsa Sadigh and
                  Shuran Song and
                  Jiajun Wu and
                  Michael C. Yip and
                  Yuke Zhu and
                  Thomas Kollar and
                  Sergey Levine and
                  Chelsea Finn},
  editor       = {Dana Kulic and
                  Gentiane Venture and
                  Kostas E. Bekris and
                  Enrique Coronado},
  title        = {{DROID:} {A} Large-Scale In-The-Wild Robot Manipulation Dataset},
  booktitle    = {Robotics: Science and Systems XX, Delft, The Netherlands, July 15-19,
                  2024},
  year         = {2024},
  url          = {https://doi.org/10.15607/RSS.2024.XX.120},
  doi          = {10.15607/RSS.2024.XX.120},
  bibsource    = {dblp computer science bibliography, https://dblp.org}
}

@inproceedings{DBLP:conf/rss/ZhaoKLF23,
  author       = {Tony Z. Zhao and
                  Vikash Kumar and
                  Sergey Levine and
                  Chelsea Finn},
  editor       = {Kostas E. Bekris and
                  Kris Hauser and
                  Sylvia L. Herbert and
                  Jingjin Yu},
  title        = {Learning Fine-Grained Bimanual Manipulation with Low-Cost Hardware},
  booktitle    = {Robotics: Science and Systems XIX, Daegu, Republic of Korea, July
                  10-14, 2023},
  year         = {2023},
  url          = {https://doi.org/10.15607/RSS.2023.XIX.016},
  doi          = {10.15607/RSS.2023.XIX.016},
  bibsource    = {dblp computer science bibliography, https://dblp.org}
}

@article{DBLP:journals/ijrr/ChiXFCDBTS25,
  author       = {Cheng Chi and
                  Zhenjia Xu and
                  Siyuan Feng and
                  Eric Cousineau and
                  Yilun Du and
                  Benjamin Burchfiel and
                  Russ Tedrake and
                  Shuran Song},
  title        = {Diffusion policy: Visuomotor policy learning via action diffusion},
  journal      = {Int. J. Robotics Res.},
  volume       = {44},
  number       = {10-11},
  pages        = {1684--1704},
  year         = {2025},
  url          = {https://doi.org/10.1177/02783649241273668},
  doi          = {10.1177/02783649241273668},
  bibsource    = {dblp computer science bibliography, https://dblp.org}
}

@inproceedings{DBLP:conf/corl/DoshiWMDL24,
  author       = {Ria Doshi and
                  Homer Rich Walke and
                  Oier Mees and
                  Sudeep Dasari and
                  Sergey Levine},
  editor       = {Pulkit Agrawal and
                  Oliver Kroemer and
                  Wolfram Burgard},
  title        = {Scaling Cross-Embodied Learning: One Policy for Manipulation, Navigation,
                  Locomotion and Aviation},
  booktitle    = {Conference on Robot Learning, 6-9 November 2024, Munich, Germany},
  series       = {Proceedings of Machine Learning Research},
  pages        = {496--512},
  publisher    = {{PMLR}},
  year         = {2024},
  url          = {https://proceedings.mlr.press/v270/doshi25a.html},
  bibsource    = {dblp computer science bibliography, https://dblp.org}
}

@inproceedings{DBLP:conf/corl/ZawalskiCPMFL24,
  author       = {Michal Zawalski and
                  William Chen and
                  Karl Pertsch and
                  Oier Mees and
                  Chelsea Finn and
                  Sergey Levine},
  editor       = {Pulkit Agrawal and
                  Oliver Kroemer and
                  Wolfram Burgard},
  title        = {Robotic Control via Embodied Chain-of-Thought Reasoning},
  booktitle    = {Conference on Robot Learning, 6-9 November 2024, Munich, Germany},
  series       = {Proceedings of Machine Learning Research},
  pages        = {3157--3181},
  publisher    = {{PMLR}},
  year         = {2024},
  url          = {https://proceedings.mlr.press/v270/zawalski25a.html},
  bibsource    = {dblp computer science bibliography, https://dblp.org}
}

\clearpage

\clearpage
\appendix
\startcontents[chapters]
\setcounter{page}{1}

\begin{center}
  \textbf{\Large StableVLA: Towards Robust Vision-Language-Action Models without Extra Data} \vspace{0.5cm} \\
  {\Large Appendix}
  \vspace{0.5cm}
\end{center}

\printcontents[chapters]{}{1}{}

\section{Theoretical Derivation}
\label{app:proof_ib_sa}

In this section, we provide a rigorous derivation of Proposition \ref{prop:ib_sa_channel}. Let $\mathbf{X} = [\mathbf{c}_1, \dots, \mathbf{c}_D] \in \mathbb{R}^{N \times D}$ be the visual encoder output, where each column $\mathbf{c}_j \in \mathbb{R}^N$ represents a channel-wise feature vector across $N$ spatial tokens. Our goal is to find an optimal compressed representation $\mathbf{Z}$ that minimizes the IB Lagrangian:
\begin{equation}
    \mathcal{L} = I(\mathbf{X}; \mathbf{Z}) - \beta I(\mathbf{Z}; \mathbf{S}),
\end{equation}
where $\mathbf{S}$ is the target clean code. Following the IB framework for unsupervised clustering, we treat each channel $\mathbf{c}_j$ as a data point (indexed by $j$) to be clustered into $D$ semantic groups (indexed by $c$). We assume the data distribution follows $p(\mathbf{s}|j)=\mathcal{N}(\mathbf{s}|\mathbf{c}_j, \epsilon^2\mathbf{I})$ where $\epsilon$ is a smoothing parameter. The $t$-th iterative step for the soft assignment of channel $j$ to cluster $c$ is given by~\citep{tishby2000information}:
\begin{align}
    \label{eq:ib_iteration}
    q^{(t)}(c|j) &= \frac{p^{(t-1)}(c)}{Z(\mathbf{c}_j, \beta)} \exp\left[ -\beta D_{KL}[p(\mathbf{s}|j) \| p^{(t-1)}(\mathbf{s}|c)] \right],\\
    p^{(t)}(c)&=\frac{n_c^{(t)}}{D},\notag
\end{align}
where $Z(\mathbf{c}_j, \beta)$ is the partition function. We approximate the cluster-conditional distribution $p(\mathbf{s}|c)$ with a Gaussian $g(\mathbf{s}|c) = \mathcal{N}(\mathbf{s}|\boldsymbol{\mu}_c, \Sigma_c)$ and assume $\epsilon$ is sufficiently small. The KL divergence between $p(\mathbf{s}|j)$ and $g(\mathbf{s}|c)$ possesses the closed form:
\begin{align}
    D_{KL}[p(\mathbf{s}|j) \| g(\mathbf{s}|c)] &= \frac{1}{2}\left[ \log\frac{|\Sigma_c|}{|\epsilon^2 \mathbf{I}|} - N + \mathrm{tr}(\Sigma_c^{-1} \epsilon^2 \mathbf{I}) + (\mathbf{c}_j - \boldsymbol{\mu}_c)^\top \Sigma_c^{-1} (\mathbf{c}_j - \boldsymbol{\mu}_c) \right] \notag\\
    &\xrightarrow{\epsilon\to 0} \frac{1}{2}\left[\log |\Sigma_c| + (\mathbf{c}_j - \boldsymbol{\mu}_c)^\top \Sigma_c^{-1} (\mathbf{c}_j - \boldsymbol{\mu}_c)\right] + \mathrm{const} \notag \\
    &\propto (\mathbf{c}_j - \boldsymbol{\mu}_c)^\top \Sigma_c^{-1} (\mathbf{c}_j - \boldsymbol{\mu}_c) + \log |\Sigma_c|. \label{eq:klterm}
\end{align}
Plugging \cref{eq:klterm} into \cref{eq:ib_iteration}, we obtain:
\begin{align}
    q^{(t)}(c|j) =\frac{n_c^{(t-1)}/D}{|\Sigma_c|^{\beta/2}}\frac{\exp\left(-\frac{\beta}{2} (\mathbf{c}_j - \boldsymbol{\mu}_c)^\top \Sigma^{-1} (\mathbf{c}_j - \boldsymbol{\mu}_c) \right)}{Z(\mathbf{c}_j,\beta)}, \notag
\end{align}
where the constant terms are absorbed into the $Z(\mathbf{c}_j,\beta)$. Looking into the quadratic form $(\mathbf{c}_j - \boldsymbol{\mu}_c)^\top \Sigma^{-1} (\mathbf{c}_j - \boldsymbol{\mu}_c)$, for a fixed $j$, the term $\mathbf{c}_j^\top \Sigma^{-1} \mathbf{c}_j$ is constant across all clusters $c$ and can be absorbed into $Z(\mathbf{c}_j, \beta)$. Without loss of generality, we assume a shared covariance $\Sigma_c=\Sigma$ across clusters and normalized cluster centers such that $\boldsymbol{\mu}_c^\top \Sigma^{-1} \boldsymbol{\mu}_c = 1$, we have:
\begin{align}
    q^{(t)}(c|j)=\frac{n_c^{(t-1)}/D}{|\Sigma|^{\beta/2}}\frac{\exp\left(\beta\boldsymbol{\mu}_c^\top \Sigma^{-1}\mathbf{c}_j\right)}{Z(\mathbf{c}_j, \beta)}.\notag
\end{align}
The specific form of the partition function $Z(\mathbf{c}_j, \beta)$ determines the normalization and the resulting attention mechanism. We now consider two cases based on different assumptions about the latent structure.

\paragraph{Case 1: Categorical Latent Structure.} Under the assumption that each channel $j$ must be assigned to \emph{exactly one} cluster, the assignments form a categorical distribution. The partition function enforcing this constraint is $Z(\mathbf{c}_j,\beta) = \sum_{c} \exp\left( \beta \boldsymbol{\mu}_{c}^\top \Sigma^{-1} \mathbf{c}_j \right)$, leading to:
\begin{equation}
   q^{(t)}(c|j)=\frac{n_c^{(t-1)}/D}{|\Sigma|^{\beta/2}}\frac{\exp\left(\beta\boldsymbol{\mu}_c^\top \Sigma^{-1}\mathbf{c}_j\right)}{\sum_{c} \exp\left( \beta \boldsymbol{\mu}_{c}^\top \Sigma^{-1} \mathbf{c}_j \right)}. \notag
\end{equation}
We define the output representation $\mathbf{z}_c^{(t)}$ as these updated cluster centers, i.e.
\begin{align}
    \mathbf{z}_c^{(t)} := \boldsymbol{\mu}_c^{(t)} &=\frac{1}{n_c^{(t)}}\sum_{j=1}^D q^{(t)}(c|j)\mathbf{c}_j=\sum_{j=1}^D \frac{n_c^{(t-1)}/D}{n_c^{(t)}|\Sigma|^{\beta/2}}\frac{\exp\left(\beta\boldsymbol{\mu}_c^\top \Sigma^{-1}\mathbf{c}_j\right)}{\sum_{c} \exp\left( \beta \boldsymbol{\mu}_{c}^\top \Sigma^{-1} \mathbf{c}_j \right)}\\
    &=\sum_{j=1}^D \frac{\exp\left(\beta\mathbf{k}_c^\top \mathbf{q}_j\right)}{\sum_{c} \exp\left(\beta\mathbf{k}_c^\top \mathbf{q}_j\right)}\mathbf{v}_j=\sum_{j=1}^D \text{Softmax}_c(\beta \mathbf{k}_c^\top \mathbf{q}_j)\mathbf{v}_j,
\end{align}
where we define  $\mathbf{q}_j = \Sigma^{-1} \mathbf{c}_j$,  $\mathbf{k}_c = \boldsymbol{\mu}_c^{(t-1)}$ and $\mathbf{v}_j=\frac{n_c^{(t-1)}/D}{n_c^{(t)}|\Sigma|^{\beta/2}}\mathbf{c}_j$. In matrix form, this yields,
\begin{equation}
    \mathbf{Z} = \mathbf{V} \cdot \mathrm{Softmax}\left( \beta \mathbf{Q}^\top \mathbf{K} \right),
\end{equation}
where $\mathbf{Q} = \Sigma^{-1}[\mathbf{c}_1, \dots, \mathbf{c}_D]=\mathbf{W}_Q\mathbf{X}$, $\mathbf{K} = [\boldsymbol{\mu}_1^{(t-1)}, \dots, \boldsymbol{\mu}_D^{(t-1)}]=\mathbf{W}_K\mathbf{X}$ , $\mathbf{V} =\frac{n_c^{(t-1)}/D}{n_c^{(t)}|\Sigma|^{\beta/2}}[\mathbf{c}_1, \dots, \mathbf{c}_D]=\mathbf{W}_V\mathbf{X}$, $\mathbf{Z} = [\boldsymbol{\mu}_1^{(t)}, \dots, \boldsymbol{\mu}_D^{(t)}]$, and $b$ is a learnable bias. Here, $\mathbf{W}_Q,\mathbf{W}_K,\mathbf{W}_V$ are learnable parameters.

\paragraph{Case 2: Independent Bernoulli Latent Structure.} We now relax the categorical constraint. Instead of requiring each channel to be assigned to exactly one semantic group, we assume that the association of channel $j$ with each cluster $c$ is an \emph{independent} binary decision. Let $a_{jc} \in \{0, 1\}$ be a binary latent variable where $a_{jc} = 1$ denotes that channel $j$ is associated with cluster $c$, following an independent Bernoulli distribution. For a specific pair $(j, c)$, the local partition function over the binary state space $\{0, 1\}$ is:
\begin{equation}
    Z(\mathbf{c}_j, \beta) = \exp\left( \beta \boldsymbol{\mu}_c^\top \Sigma^{-1} \mathbf{c}_j \right) + \exp(b),
\end{equation}
where $b$ is a learnable bias representing the activation threshold (the ``off'' state energy). Under this formulation, the probability that channel $j$ carries the semantic information of cluster $c$ is:
\begin{align}
    q^{(t)}(a_{jc}=1 | j) &= \frac{n_c/D}{|\Sigma|^{\beta/2}}\frac{\exp\left( \beta \boldsymbol{\mu}_c^\top \Sigma^{-1} \mathbf{c}_j \right)}{\exp\left( \beta \boldsymbol{\mu}_c^\top \Sigma^{-1} \mathbf{c}_j \right) + \exp(b)}
\end{align}
Similar to the derivation in Case 1, we define the output representation $\mathbf{z}_c^{(t)}$ as these updated cluster centers, i.e.
\begin{align}
    \mathbf{z}_c^{(t)} := \boldsymbol{\mu}_c^{(t)} &= \frac{1}{n_c^{(t)}}\sum_{j=1}^{D} q^{(t)}(a_{jc}=1|j) \cdot \mathbf{c}_j=\sum_{j=1}^{D} \frac{n_c^{(t-1)}/D}{n_c^{(t)}|\Sigma|^{\beta/2}}\frac{\exp\left( \beta \boldsymbol{\mu}_c^\top \Sigma^{-1} \mathbf{c}_j \right)}{\exp\left( \beta \boldsymbol{\mu}_c^\top \Sigma^{-1} \mathbf{c}_j \right) + \exp(b)} \mathbf{c}_j\\
   &= \sum_{j=1}^D \frac{\exp\left(\beta\mathbf{k}_c^\top \mathbf{q}_j\right)}{\exp\left(\beta\mathbf{k}_c^\top \mathbf{q}_j\right)+\exp(b)}\mathbf{v}_j=\sum_{j=1}^D \sigma(\beta\mathbf{k}_c^\top \mathbf{q}_j-b)\mathbf{v}_j,
\end{align}
where we define  $\mathbf{q}_j = \Sigma^{-1} \mathbf{c}_j$,  $\mathbf{k}_c = \boldsymbol{\mu}_c^{(t-1)}$ and $\mathbf{v}_j=\frac{n_c^{(t-1)}/D}{n_c^{(t)}|\Sigma|^{\beta/2}}\mathbf{c}_j$. Here $\sigma(\cdot)=\frac{1}{1+\exp(-x)}$ is the sigmoid activation. In matrix form, this yields,
\begin{equation}
    \mathbf{Z}^{(t)} = \mathbf{V} \cdot \sigma\left( \beta \mathbf{Q}^\top \mathbf{K} - b\mathbf{1}_D\mathbf{1}_D^\top \right).
\end{equation}
where $\mathbf{Q} = \Sigma^{-1}[\mathbf{c}_1, \dots, \mathbf{c}_D]=\mathbf{W}_Q\mathbf{X}$, $\mathbf{K} = [\boldsymbol{\mu}_1^{(t-1)}, \dots, \boldsymbol{\mu}_D^{(t-1)}]=\mathbf{W}_K\mathbf{X}$ , $\mathbf{V} =\frac{n_c^{(t-1)}/D}{n_c^{(t)}|\Sigma|^{\beta/2}}[\mathbf{c}_1, \dots, \mathbf{c}_D]=\mathbf{W}_V\mathbf{X}$, $\mathbf{Z} = [\boldsymbol{\mu}_1^{(t)}, \dots, \boldsymbol{\mu}_D^{(t)}]$, and $b$ is a learnable bias. Here, $\mathbf{W}_Q,\mathbf{W}_K,\mathbf{W}_V$ are learnable parameters. This establishes the functional form of our proposed \nameofmethod.

\paragraph{Remarks.} The core difference lies in the latent competition: the categorical structure forces channels to compete for assignments, enforcing $\sum_c q(c|j) = 1$, whereas the independent Bernoulli structure allows each channel-cluster pair to be evaluated independently. For VLA projectors, {\nameofmethod} is inherently more robust: uncorrelated noise channels exhibit low covariance with all semantic clusters, resulting in gate values near zero ($\sigma \approx 0$), thus effectively filtering nuisances without suppressing legitimate semantic signals.

\clearpage
\section{Implementation Details}
\label{app:implementation}

\subsection{VLM Pre-training (Alignment Stage)}
\label{app:pretraining}

Since StableVLA introduces the Fused \nameofmethod~projector, a hybrid architecture combining a standard MLP with our covariance-based {\nameofmethod} module, the projector weights differ from standard open-source checkpoints. Therefore, prior to robotic fine-tuning, we perform a Vision-Language Alignment stage to align the visual tokens produced by Fused \nameofmethod~ with the LLM's embedding space.

We strictly adhere to the Prismatic VLMs~\citep{karamcheti2024prismatic} protocol, utilizing the LLaVA-LVIS4V-LRV dataset to ensure general-purpose visual reasoning capabilities. Table~\ref{tab:all_hyperparams} (Top) details the Pre-training configurations.

\subsection{Robotic Fine-tuning \& Baselines}

Following alignment, we fine-tune the model for robotic manipulation. As described in Sec.~\ref{sec:hybrid}, Fused \nameofmethod~ employs a dual-pathway mechanism controlled by a fusion coefficient $\lambda$ and optimized using Stochastic Pathway Dropout ($p_{\text{drop}}$).

To ensure optimal performance, we tailored hyperparameters for different benchmark suites.
Comprehensive hyperparameters for both Pre-training and Fine-tuning across all benchmarks are summarized in Table~\ref{tab:all_hyperparams}.

\textbf{Baseline Configurations.}
To ensure a fair and reproducible comparison, we align all baselines with our evaluation protocol:
\begin{itemize}
    \item \textbf{OpenVLA \citep{kim2024openvla}:} We utilize the official 7B pre-trained checkpoints with standard inference settings.
    \item \textbf{OpenVLA-OFT \citep{kim2025fine-tuning}:} We employ the officially released checkpoints.
    \item \textbf{VLA-Adapter \citep{wang2025vla-adapter}:} For LIBERO, we use official checkpoints.
    For CALVIN, we re-trained the model using the official codebase under identical configurations.
    To ensure a strong baseline, we evaluated checkpoints spanning the convergence trajectory and reported the peak performance.\footnote{Specifically, we evaluated checkpoints at different training stages, yielding scores of 3.601, 4.14, 3.628, 3.92, and 4.097. We report the best result (4.14) to represent the baseline's upper bound capability.}
    \item \textbf{OpenPi ($\pi_{0.5}$) \citep{DBLP:journals/corr/abs-2504-16054}:} We utilize the officially released model weights and follow the standard evaluation protocol provided by the authors.
\end{itemize}

\begin{table}[h!]
    \centering
    \caption{Detailed Hyperparameters for StableVLA across Pre-training and Fine-tuning stages.}
    \label{tab:all_hyperparams}
    \begin{tabular}{l|cccc|c}
        \toprule
        \multicolumn{6}{c}{\textbf{Stage I: Vision-Language Pre-training}} \\
        \midrule
        \multicolumn{1}{l|}{\textbf{Base Components}} & \multicolumn{5}{l}{LLM: Qwen2.5-0.5B \quad Vision: DINO-SigLIP (224px)} \\
        \multicolumn{1}{l|}{\textbf{Optimization}} & \multicolumn{5}{l}{Global Batch: 64 \quad LR: 2e-5 \quad Precision: BF16} \\
        \multicolumn{1}{l|}{\textbf{Fused \nameofmethod~ Params}} & \multicolumn{5}{l}{Fusion Coeff. ($\lambda$): 0.3 \quad Pathway Dropout ($p_{\text{drop}}$): 0.0} \\
        \midrule
        \midrule
        \multicolumn{6}{c}{\textbf{Stage II: Robotic Fine-tuning}} \\
        \midrule
        & \multicolumn{4}{c|}{\textbf{LIBERO Benchmark}} & \textbf{CALVIN} \\
        \cmidrule(lr){2-5} \cmidrule(lr){6-6}
        \textbf{Hyperparameter} & \textbf{Spatial} & \textbf{Goal} & \textbf{Long} & \textbf{Object} & \textbf{Benchmark} \\
        \midrule
        Global Batch Size & 64 & 128 & 128 & 64 & 64 \\
        Learning Rate & 2e-4 & 2e-4 & 2e-4 & 2e-4 & 2e-4 \\
        LoRA Rank & 64 & 64 & 64 & 64 & 64 \\
        \midrule
        \multicolumn{6}{l}{\textit{Fused \nameofmethod~ Specific Parameters (Ours)}} \\
        Fusion Coeff. ($\lambda$) & 0.3 & 0.3 & 0.3 & 0.3 & 0.3 \\
        Pathway Dropout ($p_{\text{drop}}$) & 0.3 & 0.4 & 0.0 & 0.3 & 0.3 \\
        \bottomrule
    \end{tabular}
\end{table}

\clearpage
\section{Detailed Experimental Results}
  \label{app:detailed_results}

  In this appendix, we provide comprehensive quantitative results for all methods evaluated in our experiments.

  \subsection{Full Comparison on LIBERO}

  Table~\ref{tab:app_full_libero} presents the complete comparison of all methods on the LIBERO benchmark, grouped by their training paradigm. Methods are categorized into three groups: (1) \textit{OpenX Pretrain}, which includes models pretrained on the Open X-Embodiment dataset; (2) \textit{OpenX + Web Co-train}, which additionally incorporates web data during pretraining; and (3) \textit{VLM Direct FT}, which directly fine-tunes from vision-language models without robot-specific pretraining.

  % Appendix Table: Full comparison on LIBERO benchmark (grouped by training paradigm)
% Auto-generated from CSV data
\begin{table}[h]
\centering
\setlength{\tabcolsep}{3.5pt}
\footnotesize
\caption{Full comparison on LIBERO benchmark. Methods are grouped by training paradigm. We report success rate (\%). \textbf{Bold}: best, \underline{underline}: second best.}
\label{tab:app_full_libero}
\begin{tabular}{p{1.2cm}l cccc cccc cccc cccc}
\toprule
 & & \multicolumn{4}{c}{Spatial} & \multicolumn{4}{c}{Object} & \multicolumn{4}{c}{Goal} & \multicolumn{4}{c}{Long} \\
\cmidrule(lr){3-6} \cmidrule(lr){7-10} \cmidrule(lr){11-14} \cmidrule(lr){15-18}
Training & Method & C & S3 & S4 & S5 & C & S3 & S4 & S5 & C & S3 & S4 & S5 & C & S3 & S4 & S5 \\
\midrule
\multirow{2}{1.2cm}{\raggedright\scriptsize OpenX\\Pretrain} & OpenVLA & 80.0 & 40.9 & 24.6 & 14.7 & 69.6 & 18.2 & 10.4 & 2.7 & 74.0 & 38.7 & 27.0 & 16.3 & 55.5 & 20.5 & 12.4 & 7.0 \\
 & OpenVLA-OFT & 92.6 & 89.3 & \underline{84.0} & \underline{72.1} & 98.4 & 82.5 & 69.2 & 52.8 & 96.8 & \textbf{94.5} & \underline{84.6} & \underline{70.3} & \textbf{94.4} & \textbf{77.6} & 61.9 & 40.3 \\
\midrule
\multirow{1}{1.2cm}[8pt]{\raggedright\scriptsize OpenX +\\Web Co-train} & 
\rule[-10pt]{0pt}{24pt} OpenPI0.5 & \textbf{98.4} & 88.3 & 79.0 & 62.4 & \textbf{99.4} & \textbf{97.1} & \textbf{88.4} & \textbf{76.4} & 97.2 & 87.2 & 82.5 & 64.2 & 92.0 & 76.1 & \textbf{65.6} & \textbf{47.7} \\
\midrule
\multirow{2}{1.2cm}{\raggedright\scriptsize VLM\\Direct FT} & VLA-Adapter-Pro & 96.0 & \underline{93.7} & 83.3 & 58.5 & 96.8 & 71.0 & 44.1 & 29.3 & \underline{97.4} & 79.5 & 64.7 & 47.3 & \underline{94.4} & 63.5 & 41.0 & 26.2 \\
 & StableVLA & \underline{96.2} & \textbf{94.3} & \textbf{92.1} & \textbf{82.0} & \underline{98.8} & \underline{92.4} & \underline{83.6} & \underline{70.2} & \textbf{98.0} & \underline{93.4} & \textbf{85.0} & \textbf{71.9} & 93.6 & \underline{76.3} & \underline{62.4} & \underline{45.3} \\
\bottomrule
\end{tabular}
\end{table}

  \subsection{Per-Method Detailed Results}

  The following tables provide detailed per-corruption-type results for each method. Each table reports performance across all 19 corruption types (18 for tasks without Glass Blur) at severity levels 3, 4, and 5, along with clean (uncorrupted) performance.

  \paragraph{StableVLA (Ours).} Table~\ref{tab:detail_stablevla} shows the detailed results for our method on both LIBERO and CALVIN benchmarks.

  % Detailed results for StableVLA
% Auto-generated from CSV data
\begin{table}[h]
\centering
\setlength{\tabcolsep}{5pt}
\caption{Detailed results for StableVLA on LIBERO and CALVIN benchmarks. We report success rate (\%) for LIBERO and average completed tasks for CALVIN.}
\label{tab:detail_stablevla}
\begin{tabular}{l ccc ccc ccc ccc ccc}
\toprule
 & \multicolumn{12}{c}{LIBERO} & \multicolumn{3}{c}{CALVIN} \\
\cmidrule(lr){2-13} \cmidrule(lr){14-16}
 & \multicolumn{3}{c}{Spatial} & \multicolumn{3}{c}{Object} & \multicolumn{3}{c}{Goal} & \multicolumn{3}{c}{Long} & \multicolumn{3}{c}{-} \\
\cmidrule(lr){2-4} \cmidrule(lr){5-7} \cmidrule(lr){8-10} \cmidrule(lr){11-13} \cmidrule(lr){14-16}
Corruption & S3 & S4 & S5 & S3 & S4 & S5 & S3 & S4 & S5 & S3 & S4 & S5 & S3 & S4 & S5 \\
\midrule
Clean & 96.2 & - & - & 98.8 & - & - & 98.0 & - & - & 93.6 & - & - & 4.17 & - & - \\
Gaussian Noise & 96.0 & 92.8 & 74.0 & 95.4 & 85.0 & 64.8 & 95.2 & 85.2 & 53.2 & 82.8 & 53.6 & 25.0 & 2.02 & 1.26 & 0.60 \\
Shot Noise & 95.4 & 88.0 & 71.0 & 93.2 & 80.6 & 64.0 & 96.2 & 81.4 & 61.8 & 84.8 & 63.0 & 23.4 & 1.85 & 0.82 & 0.46 \\
Impulse Noise & 94.6 & 93.2 & 80.8 & 96.0 & 83.2 & 62.8 & 96.2 & 81.8 & 57.2 & 81.2 & 55.2 & 30.2 & 2.60 & 1.32 & 0.72 \\
Speckle Noise & 94.6 & 96.6 & 90.4 & 95.6 & 93.2 & 87.2 & 97.8 & 96.0 & 86.4 & 87.2 & 84.0 & 71.8 & 2.16 & 1.49 & 0.90 \\
Gaussian Blur & 91.6 & 83.0 & 49.4 & 84.4 & 45.4 & 2.4 & 91.8 & 74.0 & 42.6 & 55.4 & 31.2 & 2.8 & 2.18 & 0.94 & 0.34 \\
Glass Blur & 91.2 & 83.6 & 55.8 & - & - & - & - & - & - & - & - & - & - & - & - \\
Defocus Blur & 89.4 & 81.6 & 65.0 & 72.8 & 46.8 & 24.8 & 85.8 & 63.0 & 47.2 & 49.6 & 27.2 & 10.6 & 1.56 & 0.64 & 0.34 \\
Motion Blur & 89.6 & 90.4 & 82.6 & 98.2 & 72.0 & 27.4 & 93.4 & 59.8 & 37.6 & 66.0 & 21.6 & 1.0 & 1.23 & 0.46 & 0.24 \\
Zoom Blur & 95.8 & 93.6 & 86.2 & 60.8 & 49.8 & 46.8 & 91.0 & 82.8 & 69.0 & 36.8 & 16.8 & 11.2 & 3.48 & 2.86 & 2.10 \\
Fog & 95.0 & 94.0 & 94.4 & 99.8 & 99.6 & 94.8 & 94.4 & 95.2 & 81.4 & 78.4 & 75.0 & 57.0 & 2.84 & 1.97 & 0.88 \\
Frost & 96.0 & 93.8 & 89.8 & 84.0 & 79.4 & 67.4 & 92.0 & 88.6 & 81.0 & 72.6 & 69.2 & 54.8 & 3.11 & 2.99 & 2.64 \\
Snow & 95.4 & 94.8 & 94.2 & 97.4 & 92.2 & 95.4 & 92.0 & 85.4 & 95.4 & 61.0 & 39.2 & 33.0 & 3.45 & 2.50 & 1.62 \\
Spatter & 96.6 & 95.0 & 95.2 & 98.2 & 98.6 & 93.4 & 93.8 & 94.4 & 91.2 & 90.8 & 94.4 & 87.2 & 4.13 & 4.04 & 3.74 \\
Contrast & 94.2 & 95.2 & 68.0 & 99.0 & 98.4 & 78.6 & 97.6 & 95.6 & 56.8 & 93.6 & 90.2 & 52.6 & 3.74 & 2.30 & 0.59 \\
Brightness & 96.4 & 96.2 & 97.0 & 98.4 & 99.0 & 98.4 & 98.0 & 99.0 & 98.0 & 92.4 & 90.2 & 90.6 & 4.04 & 3.48 & 2.71 \\
Saturate & 96.6 & 96.4 & 98.0 & 98.8 & 98.2 & 97.2 & 96.6 & 98.2 & 97.8 & 93.4 & 92.0 & 89.8 & 4.20 & 4.19 & 4.17 \\
JPEG Comp. & 96.2 & 96.6 & 94.4 & 98.6 & 98.8 & 97.4 & 96.8 & 97.4 & 96.8 & 89.4 & 88.8 & 73.8 & 2.79 & 2.42 & 1.50 \\
Pixelate & 94.6 & 95.2 & 94.4 & 96.4 & 96.4 & 95.6 & 97.6 & 96.8 & 95.6 & 92.8 & 93.0 & 84.6 & 2.60 & 2.86 & 2.69 \\
Elastic Trans. & 91.6 & 90.4 & 78.2 & 95.6 & 89.0 & 65.8 & 74.4 & 55.2 & 45.2 & 65.2 & 38.8 & 16.2 & 1.94 & 1.49 & 0.99 \\
\bottomrule
\end{tabular}
\end{table}

  \paragraph{VLA-Adapter-Pro.} Table~\ref{tab:detail_vla_adapter_pro} presents results for VLA-Adapter-Pro, the strongest baseline that shares our VLM direct fine-tuning paradigm.

  % Detailed results for VLA-Adapter-Pro
% Auto-generated from CSV data
\begin{table}[h]
\centering
\setlength{\tabcolsep}{5pt}
\caption{Detailed results for VLA-Adapter-Pro on LIBERO and CALVIN benchmarks. We report success rate (\%) for LIBERO and average completed tasks for CALVIN.}
\label{tab:detail_vla_adapter_pro}
\begin{tabular}{l ccc ccc ccc ccc ccc}
\toprule
 & \multicolumn{12}{c}{LIBERO} & \multicolumn{3}{c}{CALVIN} \\
\cmidrule(lr){2-13} \cmidrule(lr){14-16}
 & \multicolumn{3}{c}{Spatial} & \multicolumn{3}{c}{Object} & \multicolumn{3}{c}{Goal} & \multicolumn{3}{c}{Long} & \multicolumn{3}{c}{-} \\
\cmidrule(lr){2-4} \cmidrule(lr){5-7} \cmidrule(lr){8-10} \cmidrule(lr){11-13} \cmidrule(lr){14-16}
Corruption & S3 & S4 & S5 & S3 & S4 & S5 & S3 & S4 & S5 & S3 & S4 & S5 & S3 & S4 & S5 \\
\midrule
Clean & 96.0 & - & - & 96.8 & - & - & 97.4 & - & - & 94.4 & - & - & 4.14 & - & - \\
Gaussian Noise & 98.2 & 93.8 & 32.4 & 97.2 & 53.8 & 0.0 & 79.4 & 53.4 & 26.0 & 69.4 & 37.0 & 1.0 & 2.35 & 1.03 & 0.29 \\
Shot Noise & 97.8 & 84.6 & 12.8 & 93.8 & 23.4 & 0.0 & 74.6 & 49.2 & 36.6 & 76.4 & 22.8 & 4.0 & 2.06 & 0.80 & 0.21 \\
Impulse Noise & 99.0 & 93.2 & 38.0 & 98.8 & 67.8 & 0.0 & 77.4 & 56.8 & 26.4 & 72.0 & 38.6 & 2.6 & 2.92 & 1.14 & 0.37 \\
Speckle Noise & 97.8 & 97.4 & 85.0 & 97.2 & 85.4 & 40.0 & 75.2 & 67.8 & 54.6 & 83.2 & 71.2 & 28.2 & 2.59 & 1.76 & 0.90 \\
Gaussian Blur & 89.2 & 56.2 & 3.8 & 28.0 & 0.0 & 0.0 & 87.6 & 72.6 & 26.8 & 37.0 & 3.2 & 0.0 & 0.76 & 0.12 & 0.00 \\
Glass Blur & 72.6 & 41.4 & 8.0 & - & - & - & - & - & - & - & - & - & - & - & - \\
Defocus Blur & 77.8 & 35.6 & 6.6 & 5.0 & 0.0 & 0.0 & 81.6 & 60.0 & 36.4 & 19.6 & 1.4 & 0.0 & 0.41 & 0.06 & 0.00 \\
Motion Blur & 88.0 & 59.2 & 29.0 & 11.6 & 0.0 & 0.0 & 89.2 & 47.2 & 21.4 & 36.2 & 8.0 & 0.0 & 0.69 & 0.08 & 0.01 \\
Zoom Blur & 87.8 & 78.0 & 71.8 & 19.2 & 6.0 & 0.0 & 73.2 & 59.0 & 41.0 & 9.0 & 1.2 & 2.0 & 2.95 & 2.43 & 1.76 \\
Fog & 98.6 & 98.0 & 93.6 & 63.4 & 11.8 & 0.2 & 77.4 & 71.0 & 46.2 & 52.4 & 37.0 & 12.6 & 2.77 & 2.38 & 1.34 \\
Frost & 93.8 & 90.2 & 81.4 & 32.2 & 22.8 & 8.2 & 59.8 & 55.0 & 43.2 & 49.4 & 36.4 & 19.0 & 3.61 & 3.46 & 3.37 \\
Snow & 97.2 & 91.6 & 95.6 & 63.0 & 7.6 & 59.2 & 58.2 & 40.6 & 57.0 & 41.8 & 6.0 & 13.6 & 3.64 & 3.03 & 2.73 \\
Spatter & 98.4 & 96.6 & 88.0 & 97.2 & 53.4 & 6.0 & 89.2 & 77.6 & 61.0 & 82.4 & 68.2 & 36.2 & 4.05 & 3.93 & 3.23 \\
Contrast & 99.4 & 97.8 & 25.8 & 88.6 & 0.0 & 0.0 & 91.4 & 79.0 & 26.0 & 68.6 & 4.4 & 0.0 & 2.54 & 0.78 & 0.01 \\
Brightness & 98.2 & 99.4 & 97.8 & 98.8 & 98.4 & 98.6 & 89.8 & 89.6 & 88.2 & 91.2 & 91.4 & 86.0 & 4.16 & 4.12 & 3.74 \\
Saturate & 99.2 & 99.2 & 99.2 & 98.4 & 98.4 & 94.0 & 98.0 & 92.8 & 88.0 & 93.8 & 80.8 & 84.0 & 4.13 & 4.13 & 4.14 \\
JPEG Comp. & 99.0 & 98.2 & 98.6 & 98.4 & 98.8 & 98.8 & 87.4 & 84.4 & 81.6 & 90.4 & 88.8 & 81.4 & 1.80 & 1.14 & 0.53 \\
Pixelate & 98.2 & 97.2 & 94.8 & 96.2 & 91.0 & 80.0 & 97.0 & 95.0 & 88.6 & 89.6 & 85.0 & 69.4 & 2.65 & 2.01 & 1.94 \\
Elastic Trans. & 90.2 & 74.8 & 50.2 & 90.8 & 75.0 & 42.6 & 44.2 & 14.2 & 2.8 & 80.4 & 56.8 & 31.8 & 1.89 & 1.61 & 1.26 \\
\bottomrule
\end{tabular}
\end{table}

  \paragraph{OpenVLA.} Table~\ref{tab:detail_openvla} shows results for the base OpenVLA model.

  % Detailed results for OpenVLA
% Auto-generated from CSV data
\begin{table}[t]
\centering
\setlength{\tabcolsep}{6pt}
\caption{Detailed results for OpenVLA on LIBERO benchmark. We report success rate (\%).}
\label{tab:detail_openvla}
\begin{tabular}{l ccc ccc ccc ccc}
\toprule
 & \multicolumn{3}{c}{Spatial} & \multicolumn{3}{c}{Object} & \multicolumn{3}{c}{Goal} & \multicolumn{3}{c}{Long} \\
\cmidrule(lr){2-4} \cmidrule(lr){5-7} \cmidrule(lr){8-10} \cmidrule(lr){11-13}
Corruption & S3 & S4 & S5 & S3 & S4 & S5 & S3 & S4 & S5 & S3 & S4 & S5 \\
\midrule
Clean & 80.0 & - & - & 69.6 & - & - & 74.0 & - & - & 55.5 & - & - \\
Gaussian Noise & 50.2 & 0.6 & 0.0 & 0.6 & 0.0 & 0.0 & 31.4 & 4.8 & 0.0 & 8.4 & 0.2 & 0.0 \\
Shot Noise & 39.0 & 0.0 & 0.0 & 0.4 & 0.0 & 0.0 & 23.6 & 2.8 & 0.0 & 5.6 & 0.0 & 0.0 \\
Impulse Noise & 55.2 & 1.4 & 0.0 & 3.0 & 0.0 & 0.0 & 41.0 & 4.8 & 0.0 & 9.2 & 1.6 & 0.0 \\
Speckle Noise & 61.0 & 26.0 & 2.8 & 5.2 & 0.2 & 0.0 & 40.8 & 17.2 & 8.4 & 17.6 & 3.8 & 0.0 \\
Gaussian Blur & 0.0 & 0.0 & 0.0 & 0.0 & 0.0 & 0.0 & 2.4 & 0.2 & 0.0 & 0.0 & 0.0 & 0.0 \\
Glass Blur & 0.0 & 0.0 & 0.0 & - & - & - & - & - & - & - & - & - \\
Defocus Blur & 0.0 & 0.0 & 0.0 & 0.0 & 0.0 & 0.0 & 0.4 & 0.2 & 0.8 & 0.0 & 0.0 & 0.0 \\
Motion Blur & 0.0 & 0.0 & 0.0 & 0.0 & 0.0 & 0.0 & 0.6 & 0.0 & 0.0 & 0.0 & 0.0 & 0.0 \\
Zoom Blur & 7.6 & 3.2 & 0.2 & 0.6 & 0.0 & 0.0 & 14.0 & 3.4 & 1.6 & 3.0 & 0.2 & 0.0 \\
Fog & 74.6 & 69.4 & 38.0 & 46.4 & 24.2 & 0.4 & 76.2 & 69.0 & 40.2 & 51.0 & 46.4 & 20.2 \\
Frost & 20.2 & 16.8 & 4.4 & 0.6 & 0.2 & 0.4 & 28.4 & 22.8 & 14.0 & 15.6 & 12.0 & 4.2 \\
Snow & 39.0 & 10.4 & 22.2 & 0.4 & 0.0 & 0.0 & 28.6 & 19.8 & 15.0 & 14.6 & 2.2 & 3.2 \\
Spatter & 65.4 & 32.4 & 3.8 & 14.2 & 0.2 & 0.0 & 50.2 & 37.0 & 16.2 & 36.0 & 12.6 & 2.2 \\
Contrast & 71.2 & 52.6 & 6.8 & 57.4 & 28.0 & 0.0 & 75.0 & 69.2 & 16.8 & 50.2 & 38.8 & 10.0 \\
Brightness & 78.4 & 73.8 & 62.8 & 64.6 & 53.2 & 29.0 & 76.0 & 73.4 & 60.8 & 52.6 & 45.6 & 41.2 \\
Saturate & 80.8 & 81.4 & 74.2 & 65.8 & 47.8 & 12.6 & 75.2 & 75.2 & 69.8 & 55.6 & 48.6 & 40.2 \\
JPEG Comp. & 72.4 & 69.8 & 56.6 & 42.6 & 29.4 & 5.8 & 73.8 & 63.2 & 45.0 & 31.8 & 10.0 & 4.2 \\
Pixelate & 60.0 & 29.0 & 7.2 & 24.8 & 3.6 & 0.0 & 53.6 & 21.2 & 5.2 & 15.4 & 1.0 & 0.0 \\
Elastic Trans. & 2.6 & 0.2 & 0.0 & 1.0 & 0.2 & 0.0 & 5.4 & 1.8 & 0.4 & 3.2 & 1.0 & 0.0 \\
\bottomrule
\end{tabular}
\end{table}

  \paragraph{OpenVLA-OFT.} Table~\ref{tab:detail_openvla_oft} presents results for OpenVLA with orthogonal fine-tuning.

  % Detailed results for OpenVLA-OFT
% Auto-generated from CSV data
\begin{table}[t]
\centering
\setlength{\tabcolsep}{6pt}
\caption{Detailed results for OpenVLA-OFT on LIBERO benchmark. We report success rate (\%).}
\label{tab:detail_openvla_oft}
\begin{tabular}{l ccc ccc ccc ccc}
\toprule
 & \multicolumn{3}{c}{Spatial} & \multicolumn{3}{c}{Object} & \multicolumn{3}{c}{Goal} & \multicolumn{3}{c}{Long} \\
\cmidrule(lr){2-4} \cmidrule(lr){5-7} \cmidrule(lr){8-10} \cmidrule(lr){11-13}
Corruption & S3 & S4 & S5 & S3 & S4 & S5 & S3 & S4 & S5 & S3 & S4 & S5 \\
\midrule
Clean & 92.6 & - & - & 98.4 & - & - & 96.8 & - & - & 94.4 & - & - \\
Gaussian Noise & 90.4 & 89.2 & 67.8 & 88.6 & 56.4 & 21.2 & 94.4 & 87.2 & 56.0 & 74.6 & 47.6 & 13.8 \\
Shot Noise & 90.8 & 85.4 & 70.2 & 88.0 & 85.4 & 70.2 & 97.0 & 76.0 & 57.4 & 82.4 & 45.6 & 12.0 \\
Impulse Noise & 90.2 & 85.6 & 64.8 & 90.2 & 56.0 & 25.8 & 95.6 & 84.0 & 56.4 & 80.2 & 48.0 & 18.8 \\
Speckle Noise & 92.4 & 90.0 & 84.4 & 91.4 & 83.6 & 56.8 & 96.6 & 94.2 & 80.8 & 88.4 & 81.0 & 47.0 \\
Gaussian Blur & 86.8 & 80.6 & 46.6 & 53.0 & 35.0 & 4.6 & 96.4 & 84.8 & 48.0 & 78.0 & 38.4 & 6.2 \\
Glass Blur & 84.4 & 73.6 & 42.6 & - & - & - & - & - & - & - & - & - \\
Defocus Blur & 85.6 & 69.4 & 46.8 & 47.6 & 27.6 & 11.2 & 93.0 & 76.8 & 53.4 & 57.8 & 30.2 & 11.0 \\
Motion Blur & 77.6 & 55.6 & 39.4 & 57.4 & 20.6 & 3.8 & 89.6 & 51.6 & 27.8 & 42.6 & 16.0 & 4.6 \\
Zoom Blur & 82.2 & 71.8 & 62.6 & 70.8 & 44.0 & 30.8 & 87.4 & 77.6 & 69.6 & 54.8 & 38.2 & 13.4 \\
Fog & 90.4 & 87.8 & 73.0 & 94.0 & 85.6 & 59.2 & 96.0 & 92.0 & 63.0 & 81.6 & 71.0 & 37.8 \\
Frost & 91.6 & 89.0 & 83.0 & 84.0 & 80.8 & 83.0 & 90.6 & 83.8 & 70.2 & 69.2 & 61.4 & 54.2 \\
Snow & 92.4 & 89.2 & 93.0 & 88.2 & 80.2 & 61.4 & 89.6 & 64.6 & 83.0 & 61.6 & 35.2 & 29.0 \\
Spatter & 93.2 & 90.0 & 84.0 & 96.6 & 92.6 & 86.4 & 95.4 & 90.8 & 84.6 & 88.6 & 90.8 & 71.8 \\
Contrast & 91.8 & 89.4 & 69.6 & 96.0 & 90.6 & 56.8 & 98.4 & 93.2 & 66.2 & 91.0 & 84.6 & 32.4 \\
Brightness & 93.8 & 94.2 & 92.4 & 96.0 & 94.8 & 93.0 & 99.0 & 98.2 & 97.6 & 91.2 & 91.8 & 90.0 \\
Saturate & 92.8 & 92.4 & 92.8 & 92.8 & 95.4 & 94.0 & 95.4 & 95.0 & 97.6 & 91.6 & 91.8 & 85.2 \\
JPEG Comp. & 92.2 & 92.2 & 92.6 & 97.0 & 96.8 & 91.8 & 98.6 & 96.0 & 97.2 & 92.0 & 87.4 & 83.8 \\
Pixelate & 91.6 & 90.8 & 90.4 & 86.2 & 69.6 & 64.2 & 97.0 & 97.4 & 96.6 & 91.8 & 90.0 & 84.6 \\
Elastic Trans. & 86.6 & 80.2 & 74.4 & 66.6 & 51.2 & 36.0 & 90.6 & 79.8 & 60.2 & 80.0 & 65.2 & 29.6 \\
\bottomrule
\end{tabular}
\end{table}

  \paragraph{OpenPI.} Table~\ref{tab:detail_openpi} shows results for OpenPI, which leverages internet-scale co-training.

  % Detailed results for OpenPI
% Auto-generated from CSV data
\begin{table}[t]
\centering
\setlength{\tabcolsep}{6pt}
\caption{Detailed results for OpenPI on LIBERO benchmark. We report success rate (\%).}
\label{tab:detail_openpi}
\begin{tabular}{l ccc ccc ccc ccc}
\toprule
 & \multicolumn{3}{c}{Spatial} & \multicolumn{3}{c}{Object} & \multicolumn{3}{c}{Goal} & \multicolumn{3}{c}{Long} \\
\cmidrule(lr){2-4} \cmidrule(lr){5-7} \cmidrule(lr){8-10} \cmidrule(lr){11-13}
Corruption & S3 & S4 & S5 & S3 & S4 & S5 & S3 & S4 & S5 & S3 & S4 & S5 \\
\midrule
Clean & 98.4 & - & - & 99.4 & - & - & 97.2 & - & - & 92.0 & - & - \\
Gaussian Noise & 98.4 & 92.0 & 29.8 & 98.4 & 98.0 & 58.4 & 96.2 & 78.0 & 21.6 & 88.4 & 73.2 & 13.0 \\
Shot Noise & 98.0 & 75.0 & 24.2 & 98.8 & 95.8 & 53.0 & 94.0 & 79.6 & 31.8 & 89.4 & 72.6 & 29.8 \\
Impulse Noise & 98.4 & 94.8 & 36.0 & 99.4 & 98.6 & 62.0 & 95.2 & 79.4 & 28.2 & 90.0 & 73.4 & 17.6 \\
Speckle Noise & 97.6 & 98.0 & 77.8 & 98.4 & 98.6 & 96.6 & 96.6 & 92.6 & 86.6 & 87.0 & 83.6 & 79.0 \\
Gaussian Blur & 41.4 & 10.8 & 0.0 & 88.8 & 10.0 & 0.0 & 76.4 & 44.6 & 1.6 & 11.0 & 0.4 & 0.0 \\
Glass Blur & 59.8 & 39.2 & 9.0 & - & - & - & - & - & - & - & - & - \\
Defocus Blur & 43.2 & 17.6 & 6.8 & 83.0 & 32.4 & 0.8 & 69.2 & 47.2 & 11.6 & 10.6 & 0.6 & 0.0 \\
Motion Blur & 81.6 & 38.4 & 10.8 & 96.2 & 77.8 & 34.8 & 87.0 & 55.0 & 24.4 & 62.4 & 21.6 & 1.2 \\
Zoom Blur & 80.4 & 75.8 & 74.4 & 99.4 & 98.4 & 97.4 & 89.6 & 83.2 & 77.0 & 60.0 & 0.8 & 1.0 \\
Fog & 96.8 & 92.0 & 81.0 & 98.4 & 99.0 & 97.2 & 97.0 & 96.0 & 82.4 & 87.6 & 76.4 & 50.2 \\
Frost & 95.8 & 91.0 & 77.4 & 98.8 & 98.8 & 95.4 & 85.8 & 81.6 & 72.4 & 47.6 & 74.2 & 56.2 \\
Snow & 97.2 & 91.8 & 91.4 & 99.4 & 98.2 & 98.2 & 92.8 & 77.8 & 79.8 & 89.6 & 76.0 & 40.8 \\
Spatter & 99.0 & 98.0 & 96.0 & 98.6 & 99.6 & 99.4 & 96.8 & 96.8 & 93.0 & 94.0 & 93.6 & 90.0 \\
Contrast & 98.4 & 97.0 & 94.0 & 98.6 & 99.0 & 96.0 & 98.0 & 97.2 & 93.6 & 89.6 & 88.6 & 67.4 \\
Brightness & 98.2 & 98.2 & 98.0 & 98.8 & 99.2 & 99.4 & 8.2 & 97.2 & 97.0 & 92.2 & 91.8 & 93.2 \\
Saturate & 99.2 & 99.0 & 98.4 & 97.6 & 98.2 & 98.8 & 98.4 & 97.0 & 97.2 & 92.8 & 92.2 & 94.0 \\
JPEG Comp. & 96.0 & 97.2 & 96.8 & 98.4 & 99.2 & 99.0 & 96.2 & 95.0 & 95.8 & 88.6 & 88.2 & 88.2 \\
Pixelate & 99.4 & 96.8 & 88.2 & 98.0 & 90.4 & 88.0 & 97.4 & 96.4 & 80.6 & 93.6 & 83.4 & 55.8 \\
Elastic Trans. & 99.2 & 97.8 & 96.0 & 99.6 & 99.8 & 100.0 & 94.8 & 90.4 & 81.4 & 95.0 & 90.0 & 81.6 \\
\bottomrule
\end{tabular}
\end{table}

  \subsection{Ablation Study Details}

  The following tables provide detailed results for our ablation study on adapter architecture design.

  \paragraph{IB-Adapter.} Table~\ref{tab:detail_ib_adapter} shows results for IB-Adapter, which uses only the image-bridge component without feature fusion.

  % Detailed results for IB-Adapter
% Auto-generated from CSV data
\begin{table}[t]
\centering
\setlength{\tabcolsep}{5pt}
\caption{Detailed results for IB-Adapter on LIBERO and CALVIN benchmarks. We report success rate (\%) for LIBERO and average completed tasks for CALVIN.}
\label{tab:detail_ib_adapter}
\begin{tabular}{l ccc ccc ccc ccc ccc}
\toprule
 & \multicolumn{12}{c}{LIBERO} & \multicolumn{3}{c}{CALVIN} \\
\cmidrule(lr){2-13} \cmidrule(lr){14-16}
 & \multicolumn{3}{c}{Spatial} & \multicolumn{3}{c}{Object} & \multicolumn{3}{c}{Goal} & \multicolumn{3}{c}{Long} & \multicolumn{3}{c}{-} \\
\cmidrule(lr){2-4} \cmidrule(lr){5-7} \cmidrule(lr){8-10} \cmidrule(lr){11-13} \cmidrule(lr){14-16}
Corruption & S3 & S4 & S5 & S3 & S4 & S5 & S3 & S4 & S5 & S3 & S4 & S5 & S3 & S4 & S5 \\
\midrule
Clean & 97.8 & - & - & 0.97 & - & - & 97.4 & - & - & 93.8 & - & - & 1.64 & - & - \\
Gaussian Noise & 95.2 & 94.2 & 79.4 & 96.6 & 80.0 & 15.0 & 85.8 & 78.0 & 57.2 & 88.0 & 54.4 & 9.8 & 1.61 & 1.58 & 1.47 \\
Shot Noise & 96.0 & 91.4 & 76.6 & 95.4 & 67.8 & 15.0 & 86.4 & 70.6 & 62.6 & 84.8 & 56.2 & 1.7 & 1.63 & 1.61 & 1.43 \\
Impulse Noise & 97.4 & 95.0 & 84.0 & 97.4 & 86.2 & 23.4 & 89.0 & 77.0 & 55.4 & 87.0 & 54.8 & 12.4 & 1.66 & 1.60 & 1.50 \\
Speckle Noise & 96.8 & 96.2 & 91.4 & 96.6 & 97.0 & 92.2 & 84.2 & 84.2 & 77.2 & 89.4 & 80.0 & 58.4 & 1.55 & 1.54 & 1.37 \\
Gaussian Blur & 88.6 & 80.8 & 32.4 & 88.0 & 36.6 & 2.0 & 93.6 & 86.0 & 68.4 & 56.0 & 41.2 & 5.2 & 1.52 & 1.44 & 1.26 \\
Glass Blur & 89.4 & 75.4 & 43.2 & - & - & - & - & - & - & - & - & - & - & - & - \\
Defocus Blur & 88.0 & 78.6 & 45.4 & 85.2 & 35.6 & 17.4 & 89.2 & 79.6 & 68.4 & 50.6 & 37.4 & 13.6 & 1.53 & 1.46 & 1.32 \\
Motion Blur & 93.0 & 87.6 & 74.6 & 91.4 & 40.6 & 2.0 & 91.2 & 75.2 & 56.0 & 71.0 & 21.0 & 1.8 & 1.44 & 1.23 & 1.16 \\
Zoom Blur & 96.0 & 90.4 & 88.0 & 50.4 & 13.2 & 12.0 & 92.6 & 83.4 & 66.6 & 44.0 & 18.5 & 10.4 & 1.54 & 1.59 & 1.48 \\
Fog & 98.4 & 98.0 & 93.4 & 96.4 & 96.4 & 56.8 & 86.6 & 86.0 & 65.4 & 87.6 & 81.4 & 52.2 & 1.11 & 1.09 & 0.92 \\
Frost & 95.2 & 93.8 & 89.4 & 94.0 & 85.8 & 69.8 & 75.0 & 70.4 & 63.0 & 68.8 & 60.8 & 43.2 & 1.70 & 1.66 & 1.61 \\
Snow & 98.8 & 97.4 & 96.6 & 97.4 & 89.4 & 98.4 & 84.4 & 69.8 & 83.4 & 69.8 & 48.0 & 50.2 & 1.55 & 1.30 & 1.00 \\
Spatter & 97.8 & 97.8 & 98.4 & 98.0 & 97.6 & 96.8 & 93.6 & 88.0 & 79.6 & 91.8 & 92.2 & 76.8 & 1.72 & 1.58 & 1.54 \\
Contrast & 95.4 & 96.0 & 82.2 & 97.6 & 84.2 & 0.2 & 97.2 & 85.0 & 56.4 & 90.6 & 77.0 & 5.2 & 1.39 & 0.61 & 0.31 \\
Brightness & 96.6 & 97.2 & 96.4 & 97.6 & 97.2 & 96.6 & 97.4 & 97.4 & 97.4 & 90.6 & 90.8 & 91.6 & 1.44 & 1.10 & 0.91 \\
Saturate & 97.4 & 95.6 & 96.6 & 98.0 & 98.0 & 98.0 & 97.4 & 97.0 & 95.4 & 92.0 & 92.4 & 90.4 & 1.69 & 1.67 & 1.66 \\
JPEG Comp. & 96.4 & 94.4 & 89.6 & 98.2 & 98.6 & 97.6 & 95.6 & 88.0 & 90.6 & 93.0 & 90.6 & 84.6 & 1.60 & 1.55 & 1.54 \\
Pixelate & 94.6 & 94.4 & 93.6 & 98.0 & 96.0 & 95.6 & 97.8 & 95.0 & 94.0 & 91.0 & 84.4 & 78.4 & 1.57 & 1.48 & 1.58 \\
Elastic Trans. & 88.4 & 79.2 & 63.6 & 95.2 & 92.6 & 69.6 & 79.0 & 66.4 & 51.4 & 57.8 & 39.2 & 16.0 & 1.77 & 1.70 & 1.71 \\
\bottomrule
\end{tabular}
\end{table}

  \paragraph{Fused IB-Adapter-softmax.} Table~\ref{tab:detail_fusedib_adapter_softmax} presents results for a variant using softmax normalization instead of sigmoid in the fusion mechanism.

  % Detailed results for FusedIB-Adapter-softmax
% Auto-generated from CSV data
\begin{table}[t]
\centering
\setlength{\tabcolsep}{5pt}
\caption{Detailed results for Fused IB-Adapter-softmax on LIBERO and CALVIN benchmarks. We report success rate (\%) for LIBERO and average completed tasks for CALVIN.}
\label{tab:detail_fusedib_adapter_softmax}
\begin{tabular}{l ccc ccc ccc ccc ccc}
\toprule
 & \multicolumn{12}{c}{LIBERO} & \multicolumn{3}{c}{CALVIN} \\
\cmidrule(lr){2-13} \cmidrule(lr){14-16}
 & \multicolumn{3}{c}{Spatial} & \multicolumn{3}{c}{Object} & \multicolumn{3}{c}{Goal} & \multicolumn{3}{c}{Long} & \multicolumn{3}{c}{-} \\
\cmidrule(lr){2-4} \cmidrule(lr){5-7} \cmidrule(lr){8-10} \cmidrule(lr){11-13} \cmidrule(lr){14-16}
Corruption & S3 & S4 & S5 & S3 & S4 & S5 & S3 & S4 & S5 & S3 & S4 & S5 & S3 & S4 & S5 \\
\midrule
Clean & 72.2 & - & - & 96.0 & - & - & 97.2 & - & - & 93.2 & - & - & 0.46 & - & - \\
Gaussian Noise & 71.4 & 50.6 & 9.4 & 96.6 & 93.0 & 83.0 & 90.6 & 67.8 & 41.8 & 59.2 & 28.8 & 3.4 & 0.47 & 0.46 & 0.46 \\
Shot Noise & 65.0 & 34.0 & 6.0 & 95.8 & 90.2 & 85.6 & 90.4 & 66.2 & 45.6 & 60.4 & 20.6 & 2.6 & 0.45 & 0.47 & 0.45 \\
Impulse Noise & 68.6 & 50.0 & 11.8 & 97.8 & 93.4 & 87.0 & 90.6 & 66.0 & 44.8 & 62.0 & 29.2 & 4.2 & 0.47 & 0.46 & 0.46 \\
Speckle Noise & 70.2 & 55.8 & 36.8 & 97.2 & 96.2 & 90.4 & 93.2 & 89.6 & 75.0 & 71.8 & 56.2 & 29.4 & 0.47 & 0.47 & 0.46 \\
Gaussian Blur & 38.6 & 11.6 & 0.0 & 93.0 & 82.2 & 20.2 & 93.2 & 76.8 & 46.2 & 60.2 & 35.4 & 0.4 & 0.46 & 0.45 & 0.46 \\
Defocus Blur & 30.6 & 11.2 & 0.6 & 93.0 & 76.8 & 51.8 & 90.2 & 67.8 & 53.8 & 52.8 & 32.0 & 5.2 & 0.46 & 0.44 & 0.46 \\
Motion Blur & 49.0 & 21.8 & 6.8 & 93.6 & 69.0 & 20.2 & 92.8 & 59.2 & 39.0 & 56.4 & 10.4 & 0.2 & 0.46 & 0.46 & 0.45 \\
Zoom Blur & 29.4 & 19.8 & 7.8 & 48.0 & 5.8 & 18.0 & 80.8 & 73.0 & 56.0 & 28.8 & 10.0 & 1.4 & - & - & - \\
Fog & 67.0 & 55.6 & 25.6 & 97.8 & 98.6 & 98.6 & 95.2 & 93.2 & 75.4 & 81.2 & 66.8 & 17.8 & 0.47 & 0.46 & 0.46 \\
Frost & 58.4 & 52.4 & 44.8 & 85.0 & 80.8 & 68.0 & 71.8 & 65.8 & 54.4 & 34.0 & 29.4 & 13.0 & 0.44 & 0.47 & 0.45 \\
Snow & 53.0 & 47.6 & 52.2 & 96.2 & 85.0 & 94.8 & 86.2 & 66.6 & 80.6 & 58.8 & 28.0 & 20.4 & 0.46 & 0.47 & 0.45 \\
Spatter & 71.0 & 56.2 & 46.2 & 94.4 & 97.6 & 96.0 & 94.2 & 87.8 & 79.4 & 71.4 & 80.2 & 59.8 & 0.46 & 0.46 & 0.46 \\
Contrast & 73.0 & 55.6 & 6.0 & 97.6 & 98.2 & 4.2 & 97.6 & 95.6 & 64.6 & 85.0 & 68.0 & 8.0 & 0.46 & 0.46 & 0.47 \\
Brightness & 74.0 & 73.2 & 76.4 & 97.4 & 96.6 & 94.4 & 98.2 & 96.8 & 97.6 & 87.6 & 89.0 & 89.4 & 0.44 & 0.46 & 0.45 \\
Saturate & 71.8 & 70.6 & 64.6 & 98.0 & 96.6 & 96.4 & 95.6 & 97.0 & 97.2 & 91.4 & 88.6 & 84.8 & 0.46 & 0.45 & 0.46 \\
JPEG Comp. & 74.0 & 70.4 & 71.4 & 97.8 & 98.2 & 97.4 & 97.4 & 95.8 & 96.0 & 79.6 & 80.4 & 71.0 & 0.46 & 0.46 & 0.46 \\
Pixelate & 61.8 & 53.8 & 52.2 & 96.4 & 94.2 & 93.8 & 97.6 & 96.6 & 95.8 & 86.8 & 77.2 & 69.6 & 0.48 & 0.47 & 0.45 \\
Elastic Trans. & 30.6 & 19.4 & 11.8 & 94.2 & 91.6 & 68.2 & 68.0 & 40.8 & 14.4 & 47.0 & 19.8 & 0.4 & 0.45 & 0.45 & 0.44 \\
\bottomrule
\end{tabular}
\end{table}

\clearpage
\section{Radar Chart Details}
  \label{app:radar}

  Figure~\ref{fig:radar} shows normalized robustness scores across 18 corruption types. For each corruption type, scores are averaged over clean images and severity levels 3--5, then normalized such that the best method equals 1. The corruption type indices are listed in Table~\ref{tab:corruption_index}.

  \begin{table}[h]
  \centering
  \caption{Corruption type indices for radar charts.}
  \label{tab:corruption_index}
  \begin{tabular}{cl|cl}
  \toprule
  Index & Corruption Type & Index & Corruption Type \\
  \midrule
  1 & Gaussian Noise & 10 & Frost \\
  2 & Shot Noise & 11 & Snow \\
  3 & Impulse Noise & 12 & Spatter \\
  4 & Speckle Noise & 13 & Contrast \\
  5 & Gaussian Blur & 14 & Brightness \\
  6 & Defocus Blur & 15 & Saturate \\
  7 & Motion Blur & 16 & JPEG Compression \\
  8 & Zoom Blur & 17 & Pixelate \\
  9 & Fog & 18 & Elastic Transform \\
  \bottomrule
  \multicolumn{4}{l}{\small *LIBERO-Spatial includes Glass Blur as index 19.}
  \end{tabular}
  \end{table}

\section{Qualitative Results of Real-world Experiments}
\label{app:real_vis}

\begin{figure}[h]
    \centering
    \includegraphics[width=\linewidth]{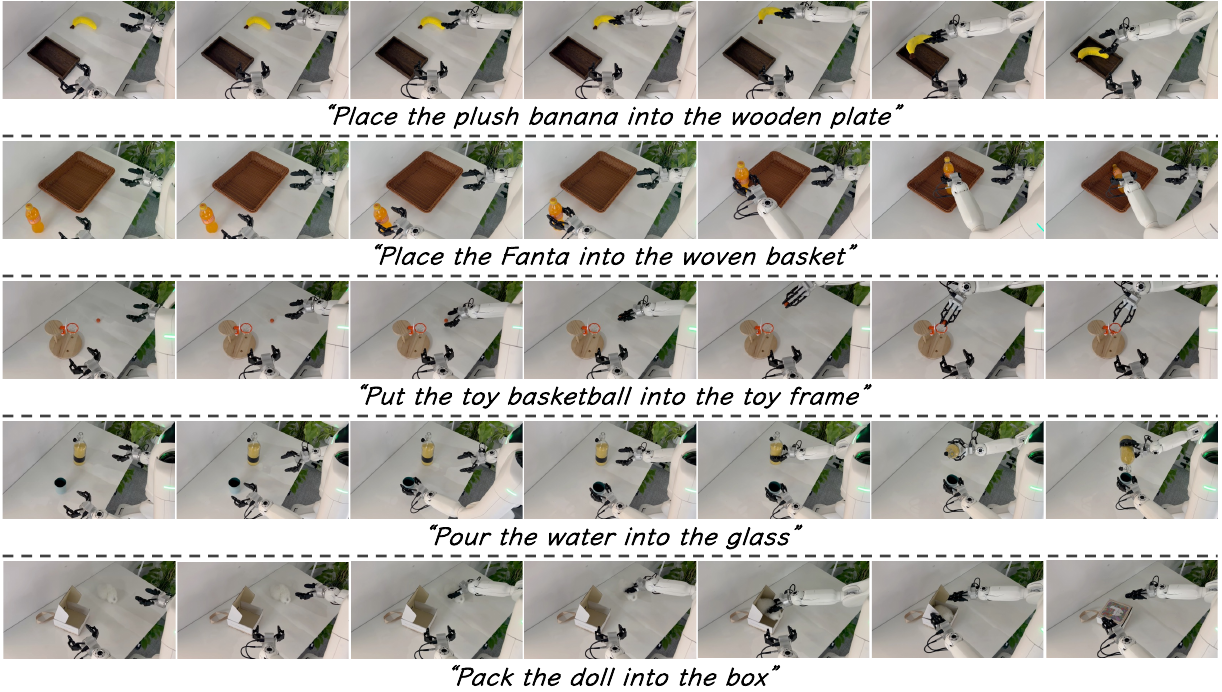}
    \caption{\textbf{Qualitative results of real-world experiments.} The figure displays successful execution sequences for five different manipulation tasks performed by the robot using StableVLA.}
    \label{fig:real_vis}
\end{figure}

\section{Detailed Related Work and Preliminaries}
\label{sec:detailed_related}

\subsection{Vision-Language-Action (VLA) Models}

Leveraging pre-trained Vision-Language Models (VLMs)~\cite{liu2023visual,comanici2025gemini,liu2024nvila,bai2025qwen2,zhu2025internvl3,xie2024show,li2024llava} for robotic control has emerged as a dominant paradigm in embodied intelligence~\citep{brohan2023rt-1, zitkovich2023rt-2, kim2024openvla, team2024octo}. However, pre-training these models from scratch relies on massive datasets, such as Open X-Embodiment~\citep{oneill2024open} and AgiBot~\cite{contributors2024agibotworldrepo}, requiring substantial computational resources. To mitigate this computational burden, VLA-Adapter~\citep{wang2025vla-adapter} proposes a resource-efficient alternative architecture. Diverging from standard paradigms, it bypasses the expensive pre-training stage on large-scale datasets, thereby directly transferring the general perceptual capabilities of VLMs to specific robotic domains.

Despite advances in training efficiency, a critical gap remains in \textit{architectural robustness}. In standard VLA models, the vision encoder~\citep{zhai2023sigmoid, oquab2023dinov2} is typically frozen during end-to-end training to preserve semantic priors~\citep{kim2024openvla, kim2025fine-tuning}, meaning input-level noise or corruption is propagated through the visual backbone. To align visual features with the downstream policy's action space, existing models use simple MLP projectors, which ideally act as the interface to suppress disturbances before they affect the policy network. Standard MLPs, while efficient at preserving spatial details, lack intrinsic mechanisms to filter out task-irrelevant nuisances. Our work addresses this by redesigning the projector's architecture to enable noise suppression during modality alignment.

\subsection{Robustness in Vision and Robotics}

Perceptual robustness is critical for the reliable deployment of robotic policies. In computer vision, this is typically evaluated using benchmarks such as ImageNet-C \cite{hendrycks2019robustness}, which introduces visual perturbations such as noise, blur, weather and digital corruptions. While classification tasks utilize the Mean Classification Error (mCE) as a standard metric, robustness in the VLA context is fundamentally tied to policy success rate under similar perturbations. Mainstream strategies to enhance robustness primarily rely on data-augmentation. In computer vision, techniques like ~\citep{hendrycks2021many} simulate corruptions during training to improve stability. Similarly, in robotic learning, Domain Randomization~\citep{tobin2017domain} is the standard approach, which randomly perturbs visual
textures or physical parameters in simulation.

However, these data-centric approaches face two significant limitations. First, they incur substantial training cost, often requiring models to be trained on vast augmented datasets. Second, these methods often rely on memorizing specific noise patterns, making it difficult to generalize to unseen corruption types. Therefore, we propose \textbf{StableVLA}, which focuses on intrinsic robustness through architectural design. We demonstrate that by reconstructing the modality alignment interface with the Information Bottleneck principle, VLA models can effectively filter visual perturbance without the need for exhaustive noise-pattern simulation.

\subsection{Attention Mechanism from the Perspective of Information Bottleneck}

Compared to CNNs, Vision Transformers (ViTs) exhibit superior robustness against various corruptions \citep{bai2021transformers, paul2022vision}. \citep{zhou2022understanding} attributes this property to the self-attention mechanism, which promotes \textit{visual grouping} where tokens aggregate into semantic clusters. This phenomenon is theoretically grounded in the Information Bottleneck (IB) principle \citep{tishby2000information}, which optimizes the trade-off between input compression and relevant information preservation. Notably, \citep{zhou2022understanding} proves that under Gaussian assumptions, the iterative optimization of the IB objective is mathematically equivalent to the self-attention operation.

While standard attention operates spatially, recent works explore visual grouping across channels. XCiT \citep{ali2021xcit} introduced Cross-Covariance Attention to compute channel-wise interactions, significantly reducing computational complexity. FAN \citep{zhou2022understanding} further establishes that this mechanism acts as subspace clustering; by applying the IB principle to the channel dimension, the model identifies coherent semantic subspaces while suppressing noisy channels. \textbf{StableVLA} extends this insight to VLA modality alignment, integrating a multi-head covariance mechanism to filter noisy channels and ensure robust semantic propagation for embodied decision-making.

\end{document}